\RequirePackage{fix-cm}
\documentclass[twocolumn]{svjour3}
\smartqed
\usepackage{graphicx}
\usepackage{xcolor}
\usepackage{multirow}
\usepackage{amsmath}
\usepackage{newtxtext,newtxmath}
\usepackage{natbib}
\usepackage{microtype}
\usepackage{stfloats}
\usepackage{balance}
\usepackage{mathrsfs}
\usepackage{textcomp}
\usepackage{manyfoot}
\usepackage{booktabs}
\usepackage{tabularx}
\usepackage{subcaption}
\usepackage[percent]{overpic}
\usepackage{enumitem}
\usepackage{tikz}
\usepackage{placeins}
\usepackage{needspace}
\usepackage{algorithm}
\usepackage{algorithmicx}
\usepackage{algpseudocode}
\usepackage{listings}
\usepackage{xurl}
\usepackage[colorlinks=true,citecolor=blue,linkcolor=red,urlcolor=blue]{hyperref}
\ExplSyntaxOn
\AddToHook{begindocument/end}{
  \NewCommandCopy{\IJCVplainref}{\ref}
  \RenewDocumentCommand{\ref}{s m}{
    \group_begin:
    \upshape
    \exp_args:Nne \regex_match:nnT
      { \A ( appendix | (sub)*section ) \. [A-Z] ( \. | \Z ) }
      { \getrefbykeydefault{#2}{anchor}{} }
      { \itshape }
    \IfBooleanTF{#1}{\IJCVplainref*{#2}}{\IJCVplainref{#2}}
    \group_end:
  }
}
\ExplSyntaxOff

\newsavebox{\bestnumbox}
\newlength{\bestdigitwidth}
\makeatletter
\newcommand{\bestnum}[2][r]{%
  \begingroup
  \sbox{\bestnumbox}{\textbf{#2}}%
  \makebox[\wd\bestnumbox][#1]{%
    \raisebox{0pt}[\ht\bestnumbox][\dp\bestnumbox]{%
      \@tfor\bestdigit:=#2\do{%
        \settowidth{\bestdigitwidth}{{\mdseries\bestdigit}}%
        \makebox[\bestdigitwidth][c]{{\fontseries{b}\selectfont\bestdigit}}%
      }%
    }%
  }%
  \endgroup
}
\makeatother

\definecolor{whatblue}{HTML}{42699E}
\definecolor{howred}{HTML}{C00000}
\renewcommand{\topfraction}{0.85}
\renewcommand{\dbltopfraction}{0.78}
\renewcommand{\textfraction}{0.18}
\journalname{ProtoFusion}
\renewcommand{\makeheadbox}{}
\hypersetup{
  pdftitle={ProtoFusion: Protocol-Guided Prompt Refinement for Generative Infrared and Visible Image Fusion},
  pdfauthor={Pengwei Liang, Junjun Jiang, Junhui Hou},
  pdfsubject={Infrared and visible image fusion},
  pdfkeywords={image fusion, generative image fusion, prompt optimization}
}

\begin{document}

\title{ProtoFusion: Protocol-Guided Prompt Refinement for Generative Infrared and Visible Image Fusion}
\titlerunning{ProtoFusion: Protocol-Guided Generative Image Fusion}
\author{Pengwei Liang \and Junjun Jiang \and Junhui Hou }
\authorrunning{Pengwei Liang et al.}
\institute{Pengwei Liang \and Junhui Hou \at
Department of Computer Science, City University of Hong Kong, Hong Kong SAR, China\\
\email{erfect2020@gmail.com; jh.hou@cityu.edu.hk}
\and
Junjun Jiang \at
School of Computer Science and Technology, Harbin Institute of Technology, Harbin, 150001, China\\
\email{jiangjunjun@hit.edu.cn}\\
Corresponding author: Junhui Hou}
\date{}
\maketitle

\begin{abstract}
Infrared and visible image fusion integrates complementary information from both modalities to provide a more comprehensive representation of the scene. We propose \textbf{ProtoFusion}, a protocol-guided generative framework for infrared and visible image fusion. Instead of relying on fixed fusion operators, hand-crafted objectives, or manually specified one-shot prompts, ProtoFusion formulates fusion as \emph{protocol-constrained generation}. It represents fusion intent with a structured protocol, consisting of environment state cards, grounding principles, and a style profile, which explicitly specifies which evidence should be preserved or suppressed and which properties the fused result should satisfy. To optimize this protocol, we introduce protocol-guided prompt refinement, a closed-loop learning framework that iteratively performs generation, evaluation, and revision under multimodal evidence. Once learned, the optimized protocol is fixed at inference time and directly guides fusion generation. To address residual source inconsistencies, we further incorporate structured belief construction, which organizes modality reliability, scene evidence, and cross-modal conflicts into a structured representation for conservative correction after generation. Extensive experiments on standard benchmarks show that ProtoFusion improves quantitative performance, achieves favorable results in aesthetic evaluation, and produces visually coherent fusion outputs. Project Page: \url{https://erfect2020.github.io/ProtoFusion/}.

\keywords{Multi-modal image fusion \and Generative image fusion \and Prompt optimization \and Structured protocol optimization}
\end{abstract}

\section{Introduction}\label{sec1}

\begin{figure*}[t]
    \centering
    \includegraphics[width=0.99\linewidth]{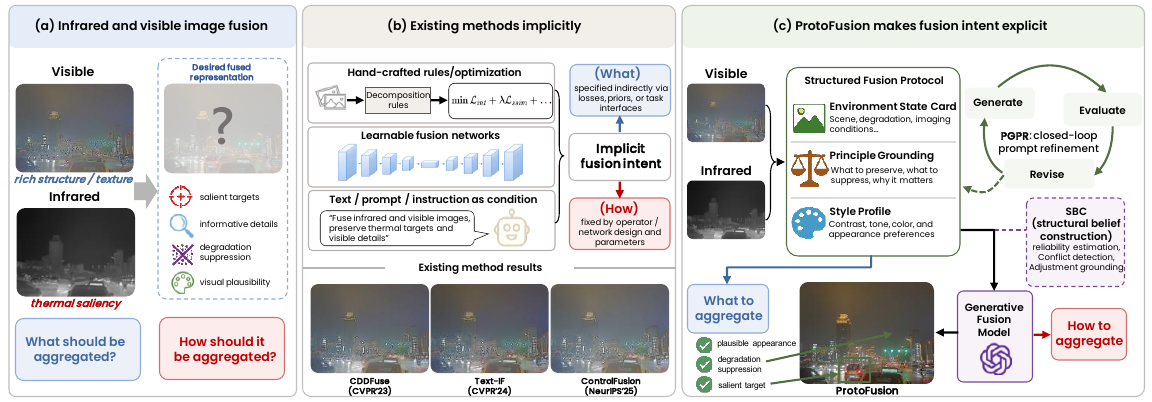}
    \caption{From implicit fusion biases to explicit protocol-constrained generation. (a) Fusion combines infrared thermal saliency with visible structure and texture. (b) Existing methods encode fusion intent implicitly through rules, networks, or conditioning. (c) ProtoFusion makes this intent explicit through an editable protocol specifying what to preserve and how to render it, with iterative refinement and optional structured belief correction.}
    \label{fig:intro}
\end{figure*}

Infrared and visible image fusion aims to combine the complementary strengths of infrared and visible imagery into a single representation that is more informative than either modality alone. As illustrated in Fig.~\ref{fig:intro}(a), infrared images provide robust thermal saliency under poor illumination and challenging visibility conditions, whereas visible images preserve richer structures and textures. Therefore, a desirable fused image should highlight thermally significant targets while retaining informative visible details for human inspection and subsequent perception in surveillance, autonomous systems, and other challenging environments~\citep{ma2019infrared,zhang2021image,zhang2023visible}. This goal naturally involves two coupled questions: what information should be aggregated from each modality, and how it should be aggregated into a coherent image. In practice, answering these questions is difficult because useful evidence must be separated from sensor noise, low-light corruption, and uninformative modality responses. This difficulty is compounded by the absence of ground-truth fused images, which forces most methods to rely on indirect objectives to define the desired fusion target~\citep{liu2026bridging,bai2025task,zhao2026mambavf}.

As summarized in Fig.~\ref{fig:intro}(b), existing methods address this problem through fixed inductive biases. Classical fusion methods rely on hand-crafted decomposition rules or explicit optimization objectives, while deep models replace them with learned networks whose fusion behavior is still largely determined by architecture design, training data, and loss functions~\citep{li2018densefuse,ma2019fusiongan,xu2020u2fusion,ma2022swinfusion,zhao2023cddfuse}. More recent methods move toward semantic-, task-, and prompt-aware fusion by incorporating detection or segmentation cues, or by conditioning the model on text and instructions~\citep{liu2022target,tang2022image,tang2023rethinking,zhao2023metafusion,liu2024task,yi2024text,liu2025promptfusion,cheng2025textfusion,zhao2024film,wang2024terf,yang2025instruction}. These designs improve adaptivity and controllability, but the underlying intent of fusion remains implicit. In particular, the \textcolor{whatblue}{\textbf{what}} is typically specified indirectly through losses, priors, or task interfaces, while the \textcolor{howred}{\textbf{how}} is constrained by fixed operators, network design, or one-shot conditioning mechanisms.  As a result, text conditioning is incorporated into the pipeline, but the fusion intent itself is rarely represented as an explicit, editable, and optimizable object.

To address this limitation, we propose \textbf{ProtoFusion}, a protocol-guided generative framework. As shown in Fig.~\ref{fig:intro}(c), ProtoFusion makes the fusion intent explicit by introducing a structured fusion protocol. The protocol specifies what evidence should be aggregated, what content should be suppressed, and what visual properties the fused result should satisfy. Under this formulation, fusion is cast as protocol-constrained generation: a multimodal generative fusion model synthesizes the fused image under the guidance of the protocol. This view is particularly attractive for modern multimodal generators, whose expressive synthesis space can accommodate target enhancement, detail recovery, degradation mitigation, and visually coherent fusion within a unified framework~\citep{zhao2023ddfm,yue2023dif,yi2024diff,shi2024vdmufusion,cao2025pdfuse}.

The proposed ProtoFusion framework uses a structured protocol with three components: environment state cards, grounding principles, and a style profile. Based on this representation, we develop \textit{protocol-guided prompt refinement} (PGPR), an offline refinement procedure that iteratively updates protocol slots through generation, evaluation, and revision under multimodal evidence. After refinement, the learned protocol is fixed and used directly to guide fusion generation at inference time. To further improve robustness on difficult cases, we additionally introduce \textit{structured belief construction} (SBC) as an optional correction module applied after PGPR. SBC analyzes residual modality imbalance and cross-modal inconsistency in the PGPR output and applies a conservative prompt-level correction when doing so is beneficial. Through this design, ProtoFusion shifts generative fusion from fixed operator design to structured protocol optimization under multimodal evidence.

Our contributions are summarized as follows.
\begin{itemize}[leftmargin=*]
    \item We propose \textbf{ProtoFusion}, a protocol-guided generative framework that represents fusion intent with a structured fusion protocol and formulates image fusion as protocol-constrained generation.
    \item We develop \textbf{PGPR}, an offline refinement procedure that optimizes the structured fusion protocol through iterative generation, evaluation, and revision under multimodal evidence. To further improve robustness, we introduce \textbf{SBC} as an optional correction module.
    \item Extensive experiments on standard infrared and visible image fusion benchmarks show that ProtoFusion achieves strong quantitative performance, consistently higher aesthetic scores, and superior visual quality under challenging visible degradations.
\end{itemize}

The remainder of this paper is organized as follows. Section~\ref{sec:related_work} reviews conventional and learning-based fusion methods, task- and language-guided approaches, generative fusion, and prompt optimization. Section~\ref{sec:method} introduces the structured fusion protocol, details the iterative optimization process of PGPR, and describes how SBC identifies and corrects residual source inconsistencies. Section~\ref{sec:experiments} presents quantitative and qualitative comparisons, evaluates downstream semantic segmentation and source consistency, and provides ablation studies, cross-generator transfer analysis, runtime evaluation, and limitations. Finally, Section~\ref{sec:conclusion} concludes the paper.

\section{Related Work}\label{sec:related_work}

\subsection{Infrared and Visible Image Fusion}

Early fusion methods rely on image decomposition and hand-designed fusion rules, while learning-based approaches use reconstruction, adversarial learning, and cross-modal feature interaction to combine infrared and visible information \citep{ma2019infrared,zhang2021image,zhang2023visible,li2018densefuse,ma2019fusiongan,xu2020u2fusion,zhao2023cddfuse}. Recent methods further model shared and modality-specific information or learn cross-modal dependencies \citep{liu2024coconet,tang2025c2rf,kang2025smlnet}. Other approaches focus on preserving complementary source information through specialized fusion mechanisms \citep{wang2024dcinn,bai2025refusion,wang2026isfuse}. Without ground-truth fused images, most methods rely on indirect objectives, such as intensity, gradient, and feature preservation. These objectives help retain source information but offer limited flexibility in specifying fusion behavior across different scenes, degradations, and user requirements.

\subsection{Task-aware, Semantic-aware, and Controllable Fusion}
Task-aware fusion uses detection and segmentation feedback to improve downstream perception \citep{liu2022target,tang2022image,tang2023rethinking,zhao2023metafusion,liu2024task}. Other methods employ discriminative learning or object-level priors to emphasize important targets and their context \citep{liu2025dcevo,zhang2026crtfs,li2025occo}. Language-guided approaches introduce more explicit control over desired content, regions, and degradation conditions \citep{yi2024text,wang2024terf,tang2025controlfusion}. Vision-language descriptions and learned semantic prompts further guide multimodal feature fusion \citep{zhao2024film,liu2025promptfusion,wang2025mtgfusion}. These methods introduce flexible semantic and task guidance, but primarily apply it through learned representations or prompt-conditioned fusion networks. More importantly, the ability of vision-language foundation models to first interpret paired sources and then formulate and refine explicit fusion instructions remains underexplored.

\subsection{Generative and Diffusion-based Image Fusion}
Diffusion models and large generative priors have recently opened a new direction for image fusion. DDFM formulates multimodal fusion with denoising diffusion models~\citep{zhao2023ddfm}, while Dif-Fusion and Diff-IF explore diffusion priors for infrared and visible image fusion and fusion-knowledge-aware generation~\citep{yue2023dif, yi2024diff}. More general diffusion-based fusion frameworks further suggest that pretrained generative models can provide a rich synthesis space for handling multiple fusion tasks and degradations~\citep{shi2024vdmufusion, cao2025pdfuse, shi2026drfusion}. Across these designs, a central issue is how to constrain generation so that perceptually plausible results retain source evidence. Preference alignment introduces another dimension of control: DPOFusion adapts a latent diffusion model to human, vision-language-model, and task-driven preferences~\citep{su2026dpofusion}. These methods demonstrate the value of generative priors for perceptual realism and restoration capability. Unlike existing generative fusion methods that mainly rely on latent optimization, sampling constraints, or model-specific guidance, our framework uses the generative model as a fusion engine and optimizes structured protocols and prompts under multimodal evidence.

\subsection{Prompt Optimization and Iterative Refinement}
Our work is also related to automatic prompt engineering and iterative self-refinement. Automatic prompt optimization improves instructions through task feedback. APE searches for effective instructions \citep{zhou2022ape}, while OPRO uses language models to iteratively optimize prompts \citep{yang2024opro}. Related approaches refine model outputs or image-editing controls: Self-Refine revises generated responses through feedback \citep{madaan2024selfrefine}, Prompt-to-Prompt controls text-driven image editing through cross-attention \citep{hertz2022prompt}, and FusionBooster analyzes and enhances initial fusion results using learned modules \citep{cheng2025fusionbooster}. Inspired by these ideas, we introduce closed-loop prompt optimization into a generative framework for infrared and visible image fusion. Unlike generic prompt optimization for language tasks or prompt editing for single-image generation, our protocol-guided refinement is grounded in paired infrared and visible observations, modality reliability, cross-modal conflicts, and fusion-specific preservation requirements. 
PGPR differs from general prompt optimization by restricting updates to structured fusion instructions and evaluating each proposal against paired source images. This allows the optimized protocol to be reused across image pairs without updating the generator.

\begin{figure*}[t]
    \centering
    \includegraphics[width=0.99\linewidth]{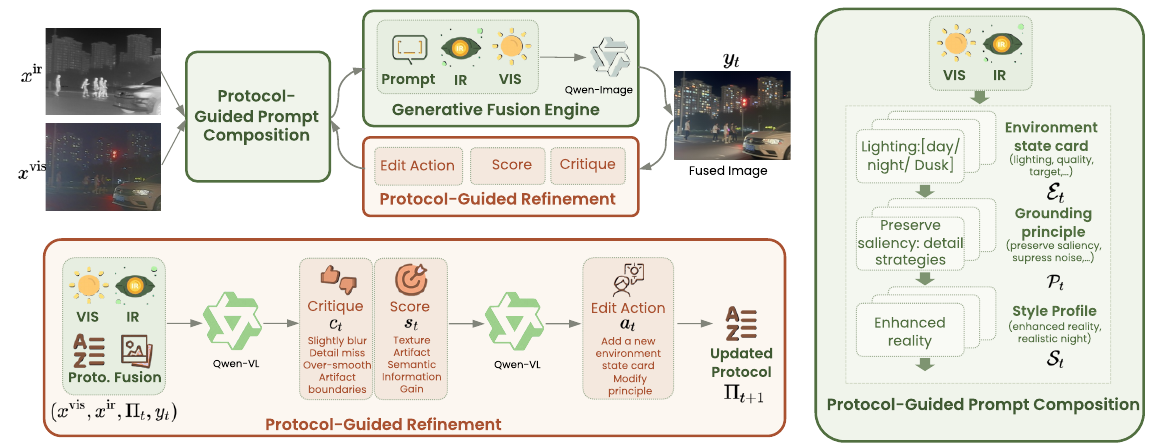}
    \caption{
        Flowchart of protocol-guided prompt refinement (PGPR). 
        PGPR is the main fusion loop. 
        At each round, it composes a prompt from a structured protocol, generates a fused image, diagnoses the current proto-fusion result, predicts a structured edit action, and updates the protocol for the next round. 
        The right panel expands protocol-guided prompt composition, where the protocol is organized into environment state cards, grounding principles, and a style profile. 
        The bottom panel expands protocol-guided refinement, where diagnosis and protocol editing are performed in two stages.
        } 
\label{fig:pgpr}
\end{figure*}

\section{Proposed Protocol-guided Generative Fusion}\label{sec:method}

\noindent\textbf{Overview.} Given a visible image $x^{\mathrm{vis}}$ and an infrared image $x^{\mathrm{ir}}$, our goal is to generate a fused image that preserves thermal saliency from the infrared modality, retains informative visible structures, and remains visually realistic. Our framework consists of a main refinement pipeline, \emph{protocol-guided prompt refinement} (PGPR), and an optional correction module applied after PGPR, \emph{structured belief construction} (SBC).

\begin{figure*}[t]
    \centering
    \includegraphics[width=0.95\linewidth]{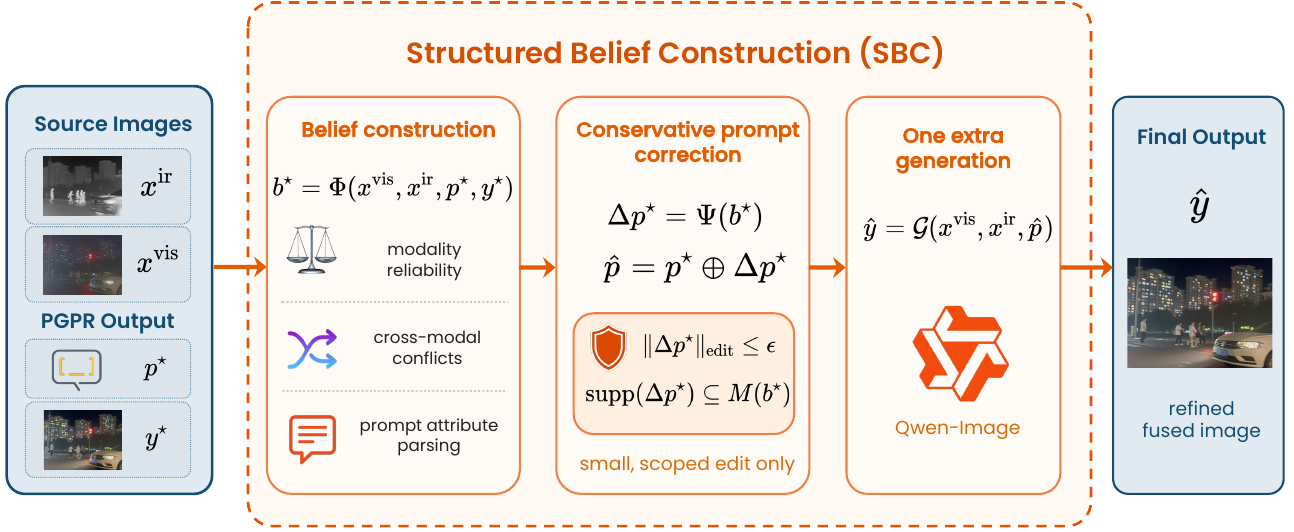}
    \caption{
        Overview of structured belief construction (SBC). Given the source images and the selected PGPR output $(p^\star, y^\star)$, SBC extracts finer-grained region-level evidence and organizes it into a belief state. This belief state captures modality reliability, cross-modal conflicts, and prompt attribute parsing. A conservative prompt correction is then predicted under a small, scoped edit constraint and used for an additional generation step.
        }
        \label{fig:sbc}
\end{figure*}

As shown in Fig.~\ref{fig:pgpr}, PGPR is the core fusion procedure. It iteratively refines a structured fusion protocol, which serves as the explicit carrier of fusion intent. At each round, the protocol is converted into a prompt, the prompt is used by the generative fusion engine to produce a fused image, and the resulting image is analyzed to update the protocol for the next round. The protocol is organized into three parts: environment state cards, grounding principles, and a style profile. Together, these components describe the source conditions, the desired fusion principles, and the target rendering style. This design makes the refinement process explicit and editable, because each round revises the protocol rather than rewriting the prompt in an unconstrained manner. After several rounds, PGPR selects the best prompt--image pair from the refinement trajectory. 
As shown in Fig.~\ref{fig:sbc}, SBC is applied only after this selection step. 
Given the source images and the selected PGPR output, SBC constructs a belief state to identify residual errors that may remain after PGPR. It then predicts a conservative prompt correction and performs one additional generation step to produce the final fused image.

\subsection{Protocol-guided Prompt Refinement}\label{sec:pgpr}

As depicted in Fig.~\ref{fig:pgpr}, PGPR contains three components: protocol-guided prompt composition, a generative fusion engine, and protocol-guided refinement. 
The following loop describes an offline refinement episode on a training pair. At round $t$, the process can be summarized as
\begin{equation}
\begin{aligned}
\Pi_t &
\xrightarrow{\mathrm{PGPC}}
p_t
\xrightarrow{\mathrm{GFE}}
y_t 
\xrightarrow{\mathrm{diag}}
(c_t, s_t)
\xrightarrow{\mathrm{edit}}
a_t 
\xrightarrow{\mathrm{update}}
\Pi_{t+1}.
\end{aligned}
\label{eq:pgpr_pipeline}
\end{equation}
Here, PGPC denotes protocol-guided prompt composition, GFE denotes the generative fusion engine, and $\mathrm{diag}$, $\mathrm{edit}$, and $\mathrm{update}$ denote source-grounded diagnosis, constrained edit prediction, and validated protocol update, respectively. The variables $\Pi_t$, $p_t$, and $y_t$ are the protocol, composed prompt, and fused candidate; $(c_t,s_t)$ are its critique and score, and $a_t$ is the proposed edit. Equation~\ref{eq:pgpr_pipeline} depicts a proposal cycle; the acceptance rule below determines which protocol is inherited.

\vspace{0.5em}
\noindent\textbf{Protocol-guided Prompt Composition.}
The protocol serves as the structured carrier of fusion intent in PGPR. At round $t$, we write it as
\begin{equation}
\Pi_t = (\mathcal{E}_t, \mathcal{P}_t, \mathcal{S}_t),
\label{eq:protocol_state}
\end{equation}
where $\mathcal{E}_t$, $\mathcal{P}_t$, and $\mathcal{S}_t$ denote environment state cards, grounding principles, and the style profile, respectively, as illustrated in Fig.~\ref{fig:pgpr}. Environment cards define which source conditions to examine, while prompt composition fills in the corresponding information from each image pair. Therefore, the protocol remains reusable, while its image-specific content adapts to each input. The initial protocol template is given in Appendix~\ref{sec:supp_protocol_prompt}. Given the source pair $(x^{\mathrm{vis}}, x^{\mathrm{ir}})$ and the current protocol $\Pi_t$, PGPC produces the prompt $p_t$ by combining image-specific information obtained according to $\mathcal{E}_t$ with the grounding principles $\mathcal{P}_t$ and the style profile $\mathcal{S}_t$.

\vspace{0.5em}
\noindent\textbf{Generative Fusion Engine.}
Given the current prompt, the generative fusion engine produces the fused image as
\begin{equation}
y_t = \mathcal{G}(x^{\mathrm{vis}}, x^{\mathrm{ir}}, p_t),
\label{eq:gfe}
\end{equation}
where $\mathcal{G}$ denotes the image generator. 
As illustrated in Fig.~\ref{fig:pgpr}, the generator takes the visible image, the infrared image, and the composed prompt as input, and outputs the current fusion result. 
This stage realizes the fusion intent encoded by the current protocol-guided prompt.

\vspace{0.5em}
\noindent\textbf{Protocol-guided Refinement.}
After generating $y_t$, PGPR diagnoses the current result and converts the diagnosis into a structured protocol edit. The updated protocol is then used to compose the next prompt. As shown in the bottom panel of Fig.~\ref{fig:pgpr}, this refinement module operates in two stages. Specifically, in the first stage, a diagnostic model analyzes the current proto-fusion result and produces a textual critique together with a scalar score:
\begin{equation}
(c_t, s_t)
=
\mathcal{R}_{\mathrm{diag}}(x^{\mathrm{vis}}, x^{\mathrm{ir}}, \Pi_t, y_t),
\label{eq:diag}
\end{equation}
where $c_t$ identifies the main deficiencies, such as missing complementary information, cross-modal inconsistencies, and visual artifacts, while $s_t$ summarizes the fusion quality for candidate selection. The critic applies predefined checks with binary judgments and confidence estimates, which are combined into a weighted score covering semantic fidelity, artifact control, texture, information gain, and protocol compliance. A fatal rejection rule excludes results containing content unsupported by either source image, regardless of their scores. The remaining candidates are ranked by their scores. A strict pass additionally requires satisfying all checks and reaching a predefined score threshold, but does not automatically terminate refinement. Although these constraints reduce source inconsistencies, the critic may still miss violations.

In the second stage, the refinement module converts the diagnosis into a structured edit action:
\begin{equation}
a_t
=
\mathcal{R}_{\mathrm{edit}}(x^{\mathrm{vis}}, x^{\mathrm{ir}}, \Pi_t, y_t, c_t, s_t),
\label{eq:edit_action}
\end{equation}
where $a_t$ is restricted to one of three actions: add an environment-state item, revise an existing principle, or add one new principle. The style profile is not edited. A deterministic validator rejects malformed, empty, repeated, or ineffective edits, as well as edits with invalid conditions. For an accepted edit, only the specified slot changes; all other slots are inherited unchanged.

The protocol is then updated by
\begin{equation}
\Pi_{t+1} = \mathcal{U}(\Pi_t, a_t),
\label{eq:update_protocol}
\end{equation}
where $\mathcal{U}(\cdot)$ applies a validated edit. The resulting proposal is generated and diagnosed before it can replace the incumbent. Invalid edits leave the incumbent unchanged.

\vspace{0.5em}
\noindent\textbf{Best-so-far Selection.}
Let $\mathcal{J}_T$ be the set of candidates actually generated and evaluated under a refinement budget $T$. The selected index is $t^\star=\arg\max_{\{t\in\mathcal{J}_T,\;s_t\ne-1\}}\{s_t\}$, with $p^\star=p_{\{t^\star\}}$ and $y^\star=y_{\{t^\star\}}$. A worse or fatal proposal is discarded, and the next edit starts from the best non-fatal state found so far. Candidate scores may rise or fall; only the retained critic score is non-decreasing. This property does not imply monotonic improvement in true image quality. The selected pair is passed to SBC when correction is enabled. The framework can use a black-box generator with the required paired-image and prompt interface. Appendix~\ref{sec:diagnostic_rules} summarizes the refinement procedure and the remaining implementation specification.

\subsection{Structured Belief Construction}
SBC is an optional correction module, as shown in Fig.~\ref{fig:sbc}. It receives the prompt--image pair to be corrected, denoted $(p^\star,y^\star)$ below. This is the selected candidate when inspecting a refinement episode, or the output generated with the fixed learned protocol at test time. SBC applies at most one conservative prompt correction and does not restart PGPR.

\vspace{0.5em}
\noindent\textbf{Belief Construction.}
Given the source images and the selected PGPR output $(p^\star, y^\star)$, SBC first constructs a belief state
\begin{equation}
b^\star
=
\Phi(x^{\mathrm{vis}}, x^{\mathrm{ir}}, p^\star, y^\star),
\label{eq:sbc_state}
\end{equation}
where $\Phi(\cdot)$ extracts finer-grained region-level evidence from the source images and the selected PGPR result. In this way, SBC moves from the coarse protocol-level evidence used in PGPR to a more localized evidence representation for correction. As shown in Fig.~\ref{fig:sbc}, it captures three aspects: modality reliability, cross-modal conflicts, and prompt attribute parsing. Intuitively, it identifies which modality is underused or over-trusted, where the remaining inconsistency lies, and which prompt attributes are most likely responsible for the current error.

\vspace{0.5em}
\noindent\textbf{Conservative Prompt Correction.}
Based on the belief state, SBC predicts a conservative prompt correction:
\begin{equation}
\Delta p^\star = \Psi(b^\star),
\qquad
\hat{p} = p^\star \oplus \Delta p^\star,
\label{eq:sbc_edit}
\end{equation}
where $\hat{p}$ is the corrected prompt. This correction is intentionally local. SBC applies a small and well-scoped modification to the selected PGPR prompt, rather than regenerating a new prompt from scratch. This constraint is written as
\begin{equation}
\|\Delta p^\star\|_{\mathrm{edit}} \le \epsilon,
\qquad
\mathrm{supp}(\Delta p^\star) \subseteq M(b^\star),
\label{eq:sbc_bound}
\end{equation}
where $\epsilon$ is a small edit budget and $M(b^\star)$ denotes the prompt-edit scope inferred from the belief state.

\vspace{0.5em}
\noindent\textbf{One Extra Generation.}
Using the corrected prompt, SBC performs one additional generation step to obtain $\hat{y} = \mathcal{G}(x^{\mathrm{vis}}, x^{\mathrm{ir}}, \hat{p})$, where $\hat{y}$ denotes the refined fused image.
In the absence of a valid correction, the selected PGPR output is retained as the final result, \emph{i.e.}, $\hat{y} = y^\star$. The correction is applied once after the PGPR selection stage.

\section{Experiments}\label{sec:experiments}

\begin{table*}[t]
\centering

\caption{
Quantitative comparison on FMB (top), MSRS (middle), and RoadScene (bottom). ProtoFusion uses PGPR+SBC on all three datasets.
Best and second-best results are highlighted in bold and underlined, respectively.
}

\label{tab:main_three_datasets}
\scriptsize
\setlength{\tabcolsep}{3.0pt}
\renewcommand{\arraystretch}{1.05}

\begin{subtable}{\textwidth}
\centering
\textbf{FMB}\par\smallskip
\resizebox{\textwidth}{!}{
\begin{tabular}{llccccccccc}
\toprule
Method & Venue
& CLIPIQA$\uparrow$ & MUSIQ$\uparrow$ & PAQ2PIQ$\uparrow$ & MANIQA$\uparrow$
& BRISQUE$\downarrow$ & TReS$\uparrow$ & EN$\uparrow$ & SD$\uparrow$ & SF$\uparrow$ \\
\midrule
CDDFuse   & CVPR'23     & 0.1912 & 54.6762 & 71.5954 & 0.2624 & 24.9722 & {62.7958} & \underline{6.9440} & \underline{38.5810} & 14.6180 \\
DDFM      & ICCV'23     & 0.1622 & 50.9695 & 69.5429 & 0.2801 & 30.1938 & 53.1768 & 6.8254 & 31.8910 & 8.8404 \\
TIMFusion & TPAMI'24    & 0.2200 & 51.3741 & 71.0020 & \underline{0.2909} & 35.3702 & 55.9598 & 6.5551 & 28.6934 & 10.9298 \\
Text-IF   & CVPR'24     & 0.1610 & 54.2197 & \underline{71.9926} & 0.2581 & 27.7682 & 61.4601 & 6.9249 & 36.4667 & 14.3592 \\
C2RF      & IJCV'25     & 0.1732 & 50.2157 & 70.9624 & 0.2507 & 27.9849 & 56.5410 & 6.8284 & 35.7855 & 12.0213 \\
DCEvo     & CVPR'25     & 0.1964 & 53.7623 & 71.6616 & 0.2651 & 25.9477 & 62.4378 & 6.8569 & 36.3743 & 13.8186 \\
Deno-IF   & NeurIPS'25  & \underline{0.2747} & 44.9853 & 67.7207 & 0.2700 & 42.4478 & 50.4824 & 6.0450 & 32.1501 & 9.7676 \\
LUT-Fuse  & ICCV'25     & 0.1688 & 53.0099 & 70.9631 & 0.2563 & \underline{19.2857} & 62.0295 & 6.7783 & 35.1191 & \underline{15.3193} \\
DRFusion  & CVPR'26 & 0.1456 & 53.3829 & 70.7716 & 0.2786 & 43.1575 & 57.7663 & 6.5361 & 32.8442 & 10.9689 \\
ISFuse    & CVPR'26 & 0.1726 & \underline{56.3705} & 71.4491 & 0.2677 & 21.8904 & \underline{67.5581} & 6.8184 & 34.2582 & 14.9685 \\
\textbf{ProtoFusion} & 
& \bestnum[c]{0.4189} & \bestnum[c]{64.6367} & \bestnum[c]{72.5243} & \bestnum[c]{0.3861}
& \bestnum[c]{14.7842} & \bestnum[c]{71.4666} & \bestnum[c]{7.0029} & \bestnum[c]{48.3840} & \bestnum[c]{17.0478} \\
\bottomrule
\end{tabular}}
\end{subtable}

\vspace{0.45em}

\begin{subtable}{\textwidth}
\centering
\textbf{MSRS}\par\smallskip
\resizebox{\textwidth}{!}{
\begin{tabular}{llccccccccc}
\toprule
Method & Venue
& CLIPIQA$\uparrow$ & MUSIQ$\uparrow$ & PAQ2PIQ$\uparrow$ & MANIQA$\uparrow$
& BRISQUE$\downarrow$ & TReS$\uparrow$ & EN$\uparrow$ & SD$\uparrow$ & SF$\uparrow$ \\
\midrule
CDDFuse   & CVPR'23     & 0.1557 & 36.5769 & 64.9346 & 0.1440 & 32.0331 & 32.5303 & 6.7854 & 43.9108 & 11.6324 \\
DDFM      & ICCV'23     & 0.1396 & 33.5653 & 62.3264 & 0.1433 & 33.8187 & 30.4202 & 6.3479 & 30.0831 & 7.6653 \\
TIMFusion & TPAMI'24    & 0.1267 & 33.5586 & 63.1841 & 0.1291 & 44.4031 & 30.1763 & 5.8837 & 40.5809 & 9.3600 \\
Text-IF   & CVPR'24     & 0.1593 & 36.3373 & 65.2339 & 0.1399 & 32.1248 & 31.9896 & \underline{6.7995} & \bestnum[c]{45.2395} & 11.8888 \\
C2RF      & IJCV'25     & 0.1363 & 32.9537 & 64.2354 & 0.1304 & 32.9751 & 30.3190 & 6.5330 & 40.7909 & 10.7847 \\
DCEvo     & CVPR'25     & 0.1299 & 36.1123 & 65.0804 & 0.1398 & 29.8427 & 32.1065 & 6.7200 & 43.4812 & 11.5661 \\
Deno-IF   & NeurIPS'25  & \underline{0.1822} & 32.0591 & 62.9750 & 0.1460 & 46.0174 & 28.5986 & 6.7202 & 41.5688 & 9.5035 \\
LUT-Fuse  & ICCV'25     & 0.1576 & 36.3014 & 65.2470 & 0.1361 & 28.6357 & 35.9565 & 6.6918 & 42.7811 & \underline{12.7832} \\
DRFusion  & CVPR'26 & 0.1066 & 35.9042 & 64.3925 & 0.1452 & 44.9263 & 35.0311 & 6.5544 & 41.9877 & 9.9147 \\
ISFuse    & CVPR'26 & 0.1813 & \underline{38.6243} & \underline{65.7127} & \underline{0.1550} & \underline{26.3307} & \underline{37.1416} & 6.4172 & 32.8123 & 12.0573 \\
\textbf{ProtoFusion} & 
& \bestnum[c]{0.3213} & \bestnum[c]{52.2836} & \bestnum[c]{68.0771} & \bestnum[c]{0.2759}
& \bestnum[c]{23.3807} & \bestnum[c]{60.7874} & \bestnum[c]{6.8341} & \underline{44.4990} & \bestnum[c]{14.6222} \\
\bottomrule
\end{tabular}}
\end{subtable}

\vspace{0.45em}

\begin{subtable}{\textwidth}
\centering
\textbf{RoadScene}\par\smallskip
\resizebox{\textwidth}{!}{
\begin{tabular}{llccccccccc}
\toprule
Method & Venue
& CLIPIQA$\uparrow$ & MUSIQ$\uparrow$ & PAQ2PIQ$\uparrow$ & MANIQA$\uparrow$
& BRISQUE$\downarrow$ & TReS$\uparrow$ & EN$\uparrow$ & SD$\uparrow$ & SF$\uparrow$ \\
\midrule
CDDFuse   & CVPR'23     & 0.1642 & \underline{50.4400} & \underline{68.8432} & 0.2066 & 24.8106 & 53.7425 & \bestnum[c]{7.5463} & \bestnum[c]{56.5661} & \underline{18.9072} \\
DDFM      & ICCV'23     & 0.1788 & 42.8392 & 65.1028 & 0.1808 & 29.3473 & 40.0075 & 7.2811 & 42.5732 & 10.7103 \\
TIMFusion & TPAMI'24    & 0.1666 & 35.0450 & 63.4439 & 0.1391 & 34.8574 & 35.1014 & 7.0091 & 37.9820 & 9.1101 \\
Text-IF   & CVPR'24     & \underline{0.2174} & 45.0695 & 67.8238 & 0.1771 & 26.1162 & 44.3801 & \underline{7.4537} & 50.6812 & 16.0900 \\
C2RF      & IJCV'25     & 0.1791 & 45.3342 & 67.0446 & 0.1854 & 24.1367 & 47.4413 & 7.3007 & 45.1677 & 14.4741 \\
DCEvo     & CVPR'25     & 0.1662 & 46.9107 & 67.9385 & 0.1967 & 25.6482 & 51.5769 & 7.2845 & 45.2938 & 14.9509 \\
Deno-IF   & NeurIPS'25  & 0.1789 & 37.4117 & 64.9687 & 0.1489 & 41.9599 & 36.1046 & 7.4220 & 47.5066 & 12.7819 \\
LUT-Fuse  & ICCV'25     & 0.1455 & 42.9383 & 66.7781 & 0.1645 & 24.8089 & 54.5410 & 6.9438 & 45.7622 & \bestnum[c]{19.6108} \\
DRFusion  & CVPR'26 & 0.1348 & 38.1821 & 65.4178 & 0.1587 & 38.9396 & 42.3365 & 7.0046 & 42.3950 & 10.0322 \\
ISFuse    & CVPR'26 & 0.1584 & 47.5480 & 67.9547 & \underline{0.2108} & \underline{23.3311} & \underline{56.5832} & 7.0224 & 37.8520 & 15.1578 \\
\textbf{ProtoFusion} &
& \bestnum[c]{0.3765} & \bestnum[c]{62.2376} & \bestnum[c]{73.1191} & \bestnum[c]{0.3220}
& \bestnum[c]{19.2434} & \bestnum[c]{77.0120} & 7.3201 & \underline{54.6165} & 18.8520 \\
\bottomrule
\end{tabular}}
\end{subtable}

\end{table*}

\subsection{Experimental Setup}\label{sec:setup}

\par\smallskip\noindent\textbf{Experimental Protocol.}
ProtoFusion uses a single protocol shared by all benchmarks. We refine this protocol offline on the FMB training split with PGPR. During refinement, each training image pair is processed with a refinement loop of $T=6$ rounds, while all model parameters remain frozen and only the protocol slots are updated. After refinement, the resulting protocol is fixed. For the base inference path, Qwen3-VL-8B-Instruct~\citep{bai2025qwen3} composes a prompt from the fixed protocol, and Qwen-Image-Edit~\citep{wu2025qwen} generates one fused image. PGPR runs offline, and inference uses the fixed optimized protocol. The PGPR+SBC variant additionally constructs a belief state and performs one corrective generation when a valid correction is available. In the primary Qwen-Image-Edit experiments, we use 40 image-generation sampling steps and disable sampling in the VLM-driven modules.
The ProtoFusion results in Tables~\ref{tab:main_three_datasets} and~\ref{tab:subjective_three_subtables} use PGPR+SBC on all three datasets. Results labeled PGPR in the ablations refer to the base path without SBC. The component and refinement-budget studies are evaluated on the FMB test pairs; Full PGPR and $T=6$ use the same base PGPR results as the SBC ablation. The main benchmark comparisons and downstream segmentation use the full test sets specified below.

\par\smallskip\noindent\textbf{Datasets and Compared Methods.}
We evaluate ProtoFusion on the full test sets of three widely used infrared and visible image fusion benchmarks: {FMB}~\citep{liu2023segmif}, {MSRS}~\citep{tang2022image}, and {RoadScene}~\citep{roadscene}, containing {280}, {361}, and {221} image pairs, respectively. 
We compare against representative state-of-the-art methods, including CDDFuse~\citep{zhao2023cddfuse}, DDFM~\citep{zhao2023ddfm}, TIMFusion~\citep{liu2024task}, Text-IF~\citep{yi2024text}, C2RF~\citep{tang2025c2rf}, DCEvo~\citep{liu2025dcevo}, Deno-IF~\citep{xu2025denoif}, LUT-Fuse~\citep{yi2025lutfuse}, DRFusion~\citep{shi2026drfusion}, and ISFuse~\citep{wang2026isfuse}.

\begin{table*}[t]
\centering
\caption{UniPercept evaluation on FMB, MSRS, and RoadScene. ProtoFusion uses PGPR+SBC on all three datasets. Higher is better; best and second-best results are bold and underlined, respectively.}
\label{tab:subjective_three_subtables}
\footnotesize
\setlength{\tabcolsep}{2.0pt}
\renewcommand{\arraystretch}{1.12}
\begin{tabular*}{\textwidth}{@{\extracolsep{\fill}}ll*{12}{r}@{}}
\toprule
& & \multicolumn{4}{c}{FMB} & \multicolumn{4}{c}{MSRS} & \multicolumn{4}{c}{RoadScene} \\
\cmidrule(lr){3-6}\cmidrule(lr){7-10}\cmidrule(lr){11-14}
Method & Venue & IAA & IQA & ISTA & Overall & IAA & IQA & ISTA & Overall & IAA & IQA & ISTA & Overall \\
\midrule
CDDFuse & CVPR'23 & 21.0344 & 26.1464 & 39.4317 & 28.8708 & 27.6337 & 28.9799 & 41.5625 & 32.7254 & 18.5933 & 19.3558 & 26.8066 & 21.5852 \\
DDFM & ICCV'23 & 22.0183 & 25.7826 & 38.4991 & 28.7667 & 26.6704 & 26.9214 & 38.7362 & 30.7760 & \underline{23.2981} & \underline{24.1244} & \underline{35.1307} & \underline{27.5177} \\
TIMFusion & TPAMI'24 & \underline{24.9254} & 27.8087 & 38.1953 & 30.3098 & 24.7261 & 26.4986 & 34.1963 & 28.4737 & 18.9853 & 19.1075 & 25.3255 & 21.1394 \\
Text-IF & CVPR'24 & 22.8638 & \underline{29.2772} & \underline{41.5920} & \underline{31.2443} & 26.6080 & 28.0942 & 41.1446 & 31.9489 & 18.9019 & 19.7031 & 27.0297 & 21.8782 \\
C2RF & IJCV'25 & 19.1790 & 23.4826 & 36.4092 & 26.3569 & 25.1766 & 26.2832 & 37.9217 & 29.7939 & 18.4992 & 18.5967 & 22.2783 & 19.7914 \\
DCEvo & CVPR'25 & 19.1839 & 22.9107 & 34.2859 & 25.4602 & 26.8705 & 28.1842 & 40.6603 & 31.9050 & 18.2865 & 18.4621 & 20.9072 & 19.2186 \\
Deno-IF & NeurIPS'25 & 20.9625 & 25.6165 & 37.5344 & 28.0378 & 23.4865 & 25.7680 & 36.7133 & 28.6559 & 17.6776 & 18.5082 & 22.9827 & 19.7229 \\
LUT-Fuse & ICCV'25 & 19.2929 & 23.2551 & 35.4638 & 26.0039 & 26.2642 & 27.1285 & 41.0203 & 31.4710 & 17.8954 & 18.4497 & 17.2435 & 17.8628 \\
DRFusion & CVPR'26 & 20.0379 & 23.9324 & 35.7415 & 26.5706 & 26.0350 & 27.2722 & 39.9891 & 31.0987 & 18.7291 & 18.9649 & 24.2574 & 20.6505 \\
ISFuse & CVPR'26 & 19.2496 & 21.8127 & 33.5917 & 24.8847 & \underline{35.7133} & \underline{29.7798} & \underline{43.7417} & \underline{36.4116} & 19.0317 & 18.8529 & 19.9400 & 19.2749 \\
\textbf{ProtoFusion} & & \bestnum{42.0143} & \bestnum{53.7696} & \bestnum{51.7683} & \bestnum{49.1841} & \bestnum{43.7136} & \bestnum{44.0845} & \bestnum{48.7273} & \bestnum{45.5085} & \bestnum{50.6442} & \bestnum{54.3835} & \bestnum{52.0622} & \bestnum{52.3633} \\
\bottomrule
\end{tabular*}
\end{table*}

\newcommand{\twoline}[2]{\shortstack[c]{#1\\#2}}
\newcommand{\oneline}[1]{\raisebox{0.5\baselineskip}{#1}}
\newcommand{\confshort}[2]{#1\textquotesingle{}#2}
\begin{figure*}[t]
    \centering
    
    \includegraphics[width=0.99\linewidth]{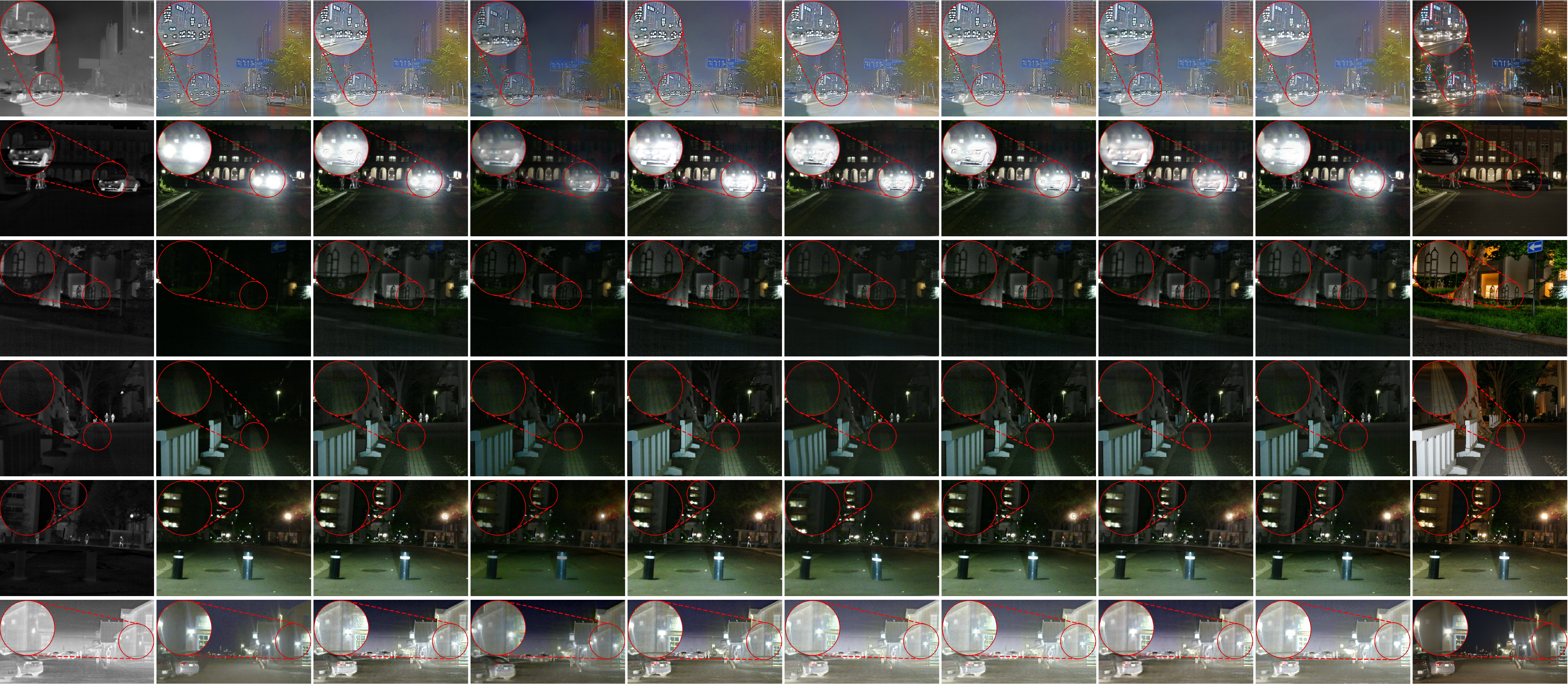}
    \fontsize{8}{9.2}\selectfont
    \setlength{\tabcolsep}{0pt}
    \renewcommand{\arraystretch}{0.95}
    \begin{tabularx}{0.99\linewidth}{
    *{10}{>{\centering\arraybackslash}X}
    }
    \oneline{Infrared}
    & \oneline{Visible}
    & \twoline{CDDFuse}{\confshort{CVPR}{23}}
    & \twoline{DDFM}{\confshort{ICCV}{23}}
    & \twoline{Text-IF}{\confshort{CVPR}{24}}
    & \twoline{C2RF}{\confshort{IJCV}{25}}
    & \twoline{DCEvo}{\confshort{CVPR}{25}}
    & \twoline{Deno-IF}{\confshort{NeurIPS}{25}}
    & \twoline{LUT-Fuse}{\confshort{ICCV}{25}}
    & \twoline{\textbf{Proto}}{\textbf{Fusion}} \\
    \end{tabularx}
    
    \caption{
    Representative visual comparison on FMB, MSRS, and RoadScene. 
    The examples illustrate target visibility and rendering under visible degradations.
    }
\label{fig:visual_compare}

\end{figure*}

\begin{table*}[tp]
\centering
\caption{
Semantic segmentation results on FMB. Best and second-best results are highlighted in bold and underlined, respectively.}
\label{tab:fmb_seg_selected}
\scriptsize
\setlength{\tabcolsep}{1pt}
\renewcommand{\arraystretch}{1.12}
\begin{tabular*}{\linewidth}{@{\extracolsep{\fill}}ll*{18}{c}@{}}
\toprule
Method & Venue & mIoU & mAcc
& \multicolumn{2}{c}{Road} & \multicolumn{2}{c}{Building}
& \multicolumn{2}{c}{Sky} & \multicolumn{2}{c}{Car}
& \multicolumn{2}{c}{Truck} & \multicolumn{2}{c}{Bus}
& \multicolumn{2}{c}{Veg.} & \multicolumn{2}{c}{Pole} \\
\cmidrule(lr){5-6}\cmidrule(lr){7-8}\cmidrule(lr){9-10}\cmidrule(lr){11-12}
\cmidrule(lr){13-14}\cmidrule(lr){15-16}\cmidrule(lr){17-18}\cmidrule(lr){19-20}
& & & & IoU & Acc & IoU & Acc & IoU & Acc & IoU & Acc
& IoU & Acc & IoU & Acc & IoU & Acc & IoU & Acc \\
\midrule
CDDFuse & CVPR'23 & 0.510 & 0.632 & 0.560 & \underline{0.811} & 0.575 & 0.700 & 0.679 & 0.692 & 0.631 & 0.884 & 0.450 & 0.499 & 0.622 & 0.837 & 0.465 & 0.474 & 0.100 & \underline{0.160} \\
DDFM & ICCV'23 & 0.472 & 0.588 & 0.536 & 0.724 & \underline{0.585} & \underline{0.709} & \underline{0.692} & \underline{0.704} & 0.555 & 0.786 & 0.269 & 0.343 & 0.565 & 0.797 & \underline{0.485} & \underline{0.497} & 0.089 & 0.144 \\
TIMFusion & TPAMI'24 & 0.511 & 0.627 & 0.560 & 0.798 & 0.555 & 0.671 & 0.607 & 0.621 & 0.633 & 0.896 & \underline{0.487} & \underline{0.551} & \underline{0.676} & 0.839 & 0.472 & 0.482 & 0.100 & 0.157 \\
Text-IF & CVPR'24 & \underline{0.513} & \underline{0.637} & \underline{0.565} & 0.807 & 0.579 & 0.706 & 0.675 & 0.687 & \underline{0.641} & 0.888 & 0.439 & 0.518 & 0.643 & \underline{0.864} & 0.471 & 0.481 & 0.095 & 0.147 \\
C2RF & IJCV'25 & 0.482 & 0.605 & 0.534 & 0.781 & 0.573 & 0.703 & 0.607 & 0.615 & 0.607 & 0.865 & 0.410 & 0.485 & 0.607 & 0.820 & 0.443 & 0.450 & 0.080 & 0.124 \\
DCEvo & CVPR'25 & 0.492 & 0.623 & 0.550 & 0.806 & 0.568 & 0.698 & 0.688 & 0.702 & 0.628 & 0.888 & 0.385 & 0.475 & 0.567 & 0.793 & 0.454 & 0.463 & 0.099 & 0.157 \\
Deno-IF & NeurIPS'25 & 0.432 & 0.524 & 0.486 & 0.663 & 0.484 & 0.572 & 0.558 & 0.566 & 0.610 & 0.831 & 0.424 & 0.466 & 0.392 & 0.532 & 0.410 & 0.416 & 0.092 & 0.146 \\
LUT-Fuse & ICCV'25 & 0.491 & 0.615 & 0.512 & 0.756 & 0.566 & 0.695 & 0.667 & 0.679 & 0.605 & 0.882 & 0.379 & 0.414 & 0.621 & 0.852 & 0.476 & 0.487 & 0.100 & 0.157 \\
DRFusion & CVPR'26 & 0.458 & 0.591 & 0.530 & 0.790 & 0.555 & 0.681 & 0.658 & 0.678 & 0.604 & \underline{0.902} & 0.283 & 0.351 & 0.505 & 0.730 & 0.423 & 0.431 & \bestnum[c]{0.105} & \bestnum[c]{0.169} \\
ISFuse & CVPR'26 & 0.507 & 0.618 & 0.538 & 0.769 & 0.577 & 0.703 & 0.634 & 0.641 & 0.633 & 0.875 & 0.456 & 0.503 & 0.668 & 0.838 & 0.450 & 0.459 & 0.096 & 0.152 \\
\midrule
\textbf{ProtoFusion} &  & \bestnum[c]{0.554} & \bestnum[c]{0.684} & \bestnum[c]{0.610} & \bestnum[c]{0.842} & \bestnum[c]{0.587} & \bestnum[c]{0.726} & \bestnum[c]{0.699} & \bestnum[c]{0.711} & \bestnum[c]{0.662} & \bestnum[c]{0.923} & \bestnum[c]{0.502} & \bestnum[c]{0.605} & \bestnum[c]{0.725} & \bestnum[c]{0.938} & \bestnum[c]{0.547} & \bestnum[c]{0.562} & \underline{0.101} & \bestnum[c]{0.169} \\
\bottomrule
\end{tabular*}
\end{table*}

\begin{figure*}[ht]
    \centering
    \includegraphics[width=0.99\linewidth]{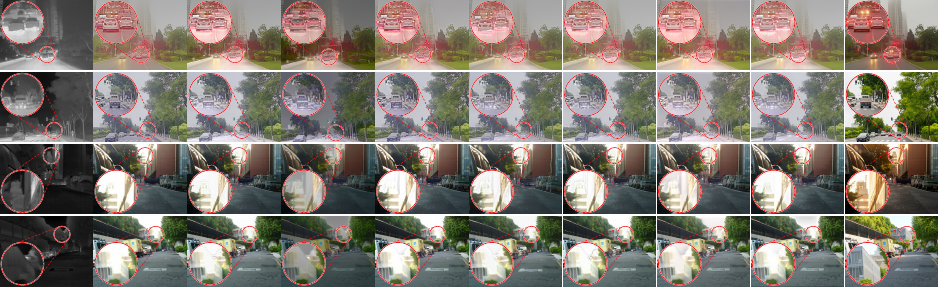}
    \fontsize{8}{9.2}\selectfont
    \setlength{\tabcolsep}{0pt}
    \renewcommand{\arraystretch}{0.95}
    \begin{tabularx}{0.99\linewidth}{*{10}{>{\centering\arraybackslash}X}}
    \oneline{Infrared}
    & \oneline{Visible}
    & \twoline{CDDFuse}{\confshort{CVPR}{23}}
    & \twoline{DDFM}{\confshort{ICCV}{23}}
    & \twoline{Text-IF}{\confshort{CVPR}{24}}
    & \twoline{C2RF}{\confshort{IJCV}{25}}
    & \twoline{DCEvo}{\confshort{CVPR}{25}}
    & \twoline{Deno-IF}{\confshort{NeurIPS}{25}}
    & \twoline{LUT-Fuse}{\confshort{ICCV}{25}}
    & \twoline{\textbf{Proto}}{\textbf{Fusion}} \\
    \end{tabularx}
    \caption{  Qualitative comparisons on degraded daytime scenes with haze and strong illumination.}
    \label{fig:visual_day}
\end{figure*}

\par\smallskip\noindent\textbf{Evaluation Metrics.}
We report six learned perceptual quality metrics, namely {CLIPIQA}~\citep{wang2023exploring}, {MUSIQ}~\citep{ke2021musiq}, {PAQ2PIQ}~\citep{ying2020patches}, {MANIQA}~\citep{yang2022maniqa}, {BRISQUE}~\citep{mittal2012brisque}, and {TReS}~\citep{golestaneh2022tres}, together with three statistical fusion metrics, {EN}, {SD}, and {SF}~\citep{wang2008performance}. In addition, we report scores from UniPercept~\citep{cao2026unipercept}, a perceptual evaluation model that assesses fused images along three complementary dimensions: image aesthetics (IAA), image quality (IQA), and structure and texture (ISTA). In addition to these three scores, we report the Overall score provided by UniPercept as its aggregate assessment. Higher values are conventionally preferred for all these metrics except {BRISQUE}, for which lower values are better. Beyond image quality, we evaluate semantic utility through downstream segmentation using mean intersection over union (mIoU) and mean accuracy (mAcc) with a frozen model. We further assess source fidelity through expert-reviewed image-level source inconsistency rates. Additional CE, $Q_{AB/F}$, and VIF results are reported in Appendix~\ref{sec:source_fidelity}.

\par\smallskip\noindent\textbf{Implementation Details.}
We instantiate the generative fusion engine $\mathcal{G}$ with Qwen-Image-Edit and use Qwen3-VL-8B-Instruct to implement the protocol-side modules, including protocol initialization $\mathcal{I}$, prompt composition $\mathcal{C}$, diagnosis $\mathcal{R}_{\mathrm{diag}}$, edit prediction $\mathcal{R}_{\mathrm{edit}}$, belief construction $\Phi$, and prompt correction $\Psi$. All model parameters remain frozen throughout both protocol refinement and inference. The Qwen-Image-Edit experiments are conducted in \texttt{PyTorch} on two BW3000 DCUs with 64\,GB of memory each. Additional details on prompt templates, model inputs, and inference settings are provided in Appendix~\ref{app:additional}.

\subsection{Performance Comparison}

\par\smallskip\noindent\textbf{Quantitative Results.}
Table~\ref{tab:main_three_datasets} compares ProtoFusion with ten competing methods on FMB, MSRS, and RoadScene. ProtoFusion achieves the best results on all six image quality metrics across the three benchmarks and leads all nine reported metrics on FMB. For the statistical metrics, Text-IF achieves the highest SD on MSRS, while CDDFuse leads EN and SD and LUT-Fuse leads SF on RoadScene. Notably, ProtoFusion maintains its advantage in perceptual quality on MSRS and RoadScene using the same protocol optimized on FMB, demonstrating its effectiveness across datasets without dataset-specific protocol refinement.

\par\smallskip\noindent\textbf{UniPercept Evaluation.}
Table~\ref{tab:subjective_three_subtables} provides a complementary perceptual evaluation using UniPercept. ProtoFusion achieves the highest IAA, IQA, ISTA, and Overall scores across all three datasets. Its Overall scores reach 49.18, 45.51, and 52.36 on FMB, MSRS, and RoadScene, outperforming the strongest competing methods by 17.94, 9.10, and 24.85 points, respectively. These consistent gains across all three evaluation dimensions further support the perceptual quality of the generated fusion results.

\par\smallskip\noindent\textbf{Qualitative Results.} 
As shown in Fig.~\ref{fig:visual_compare}, qualitative comparisons cover challenging visible conditions, including overexposure, low illumination, and noise. Across these scenes, ProtoFusion deliberately renders the fused images in a natural visible-light style, as specified by its structured protocol. This design allows infrared information to enhance salient targets and reveal obscured structures while maintaining a coherent visual appearance. Compared with competing methods that often retain visible degradations or produce indistinct target boundaries, ProtoFusion renders clearer targets, suppresses noise, and improves the overall scene appearance. Additional nighttime comparisons are provided in Appendix~\ref{sec:supp_night}.

We further examine daytime scenes with small vehicles and strong backlighting, as shown in Fig.~\ref{fig:visual_day}. In the first two rows, several competing methods introduce bright artifacts or doubled contours around vehicles, whereas ProtoFusion produces clearer vehicle boundaries and better separation from the road. In the last two rows, strong illumination obscures building structures in the visible inputs, while ProtoFusion renders more distinct rooflines and wall boundaries with a natural appearance. These comparisons highlight the central advantage of ProtoFusion: its optimized protocol explicitly controls what information to preserve or suppress from each modality and how the fused result should appear. This enables the generative model to exploit complementary infrared information and mitigate visible degradations while maintaining the intended visible-light appearance.

\subsection{Downstream Semantic Segmentation}\label{sec:segmentation}

We evaluate downstream semantic segmentation on the FMB test pairs using the same frozen CLIPer model~\citep{sun2025cliper} for each fusion method. FMB category names serve as text prompts, and the predicted segmentation maps are evaluated against pixel-level annotations.

Table~\ref{tab:fmb_seg_selected} presents the results against ten competing methods. ProtoFusion achieves an mIoU of 0.554 and an mAcc of 0.684, improving upon the strongest competitor, Text-IF, by 4.1 and 4.7 percentage points, respectively. ProtoFusion also achieves the highest IoU in seven of the eight reported classes, including car, truck, and bus. These results demonstrate improved semantic segmentation performance under the same frozen evaluation model. A key design choice helps explain these gains: ProtoFusion deliberately renders fused images with a natural visible-light appearance. This design preserves familiar visible colors, textures, and scene structures while using infrared information to enhance salient targets and obscured details. The resulting images are better suited to the visual representations learned by the frozen CLIPer model, helping explain the improvements across both object and scene segmentation.

\subsection{Hallucination Study}\label{sec:hallucination}

We evaluate image-level source inconsistencies on all FMB test pairs through a two-stage assessment. First, GPT-6 Astra examines each fused image alongside its infrared and visible sources to identify potential inconsistencies. Domain experts then manually review all fused results, including those initially judged error-free by the model, and determine the final error labels. All reported error rates are based on this comprehensive human review.

\par\smallskip\noindent\textbf{Error Categories.}
We consider four categories of source inconsistencies. Content alteration includes unsupported additions or changes to source-supported semantic attributes, text, symbols, or states. Geometric distortion covers erroneous shape, count, duplication, or spatial arrangement beyond source registration differences. Information omission denotes the loss of meaningful source-supported objects or structures. Appearance artifacts include newly introduced or amplified halos, ghosting, noise, smearing, or unnatural rendering. The categories may overlap; \emph{Any error} denotes their image-level union.

As shown in Tab.~\ref{tab:fmb-four-category-errors}, PGPR reduces the image-level \emph{Any error} rate from 53.93\% to 38.21\%, while SBC further lowers it to 19.64\%. PGPR mitigates recurring generation errors by optimizing a structured fusion protocol through source-grounded diagnosis and constrained edits, improving control over modality contributions, structural preservation, and visual appearance. SBC complements this shared protocol with region-level analysis of modality reliability and cross-modal conflicts, enabling conservative prompt corrections tailored to residual errors in each image. Together, these mechanisms reduce appearance artifacts from 25.36\% to 3.57\%, geometric distortion from 5.00\% to 0.71\%, and content alteration from 36.43\% to 14.29\%.

\begin{table}[!t]
\centering
\caption{Image-level error rates (\%) on FMB; lower is better. Categories may overlap; Any error is their union. Only definite errors enter the numerator.}
\label{tab:fmb-four-category-errors}
\small
\setlength{\tabcolsep}{2pt}
\renewcommand{\arraystretch}{1.15}
\begin{tabular*}{\linewidth}{@{\extracolsep{\fill}}lrrr@{}}
\toprule
Error category & \shortstack{Human Prompt} & PGPR & \shortstack{PGPR+SBC}\\
\midrule
Content alteration & 36.43 & 26.43 & \bestnum{14.29}\\
Geometric distortion & 5.00 & 1.79 & \bestnum{0.71}\\
Information omission & 4.64 & 3.93 & \bestnum{2.86}\\
Appearance artifacts & 25.36 & 11.43 & \bestnum{3.57}\\
\midrule
Any error & 53.93 & 38.21 & \bestnum{19.64}\\
\bottomrule
\end{tabular*}
\end{table}

\par\smallskip\noindent\textbf{Residual errors.}
 The fatal gate rejects source violations identified by the critic, while undetected errors can persist in the generated results. Appendix~\ref{sec:hallucination_appendix} traces an example in which an unsupported \emph{white umbrella} interpretation propagates into the fused output.

\begin{figure*}[tb]
    \centering
    \includegraphics[width=\linewidth]{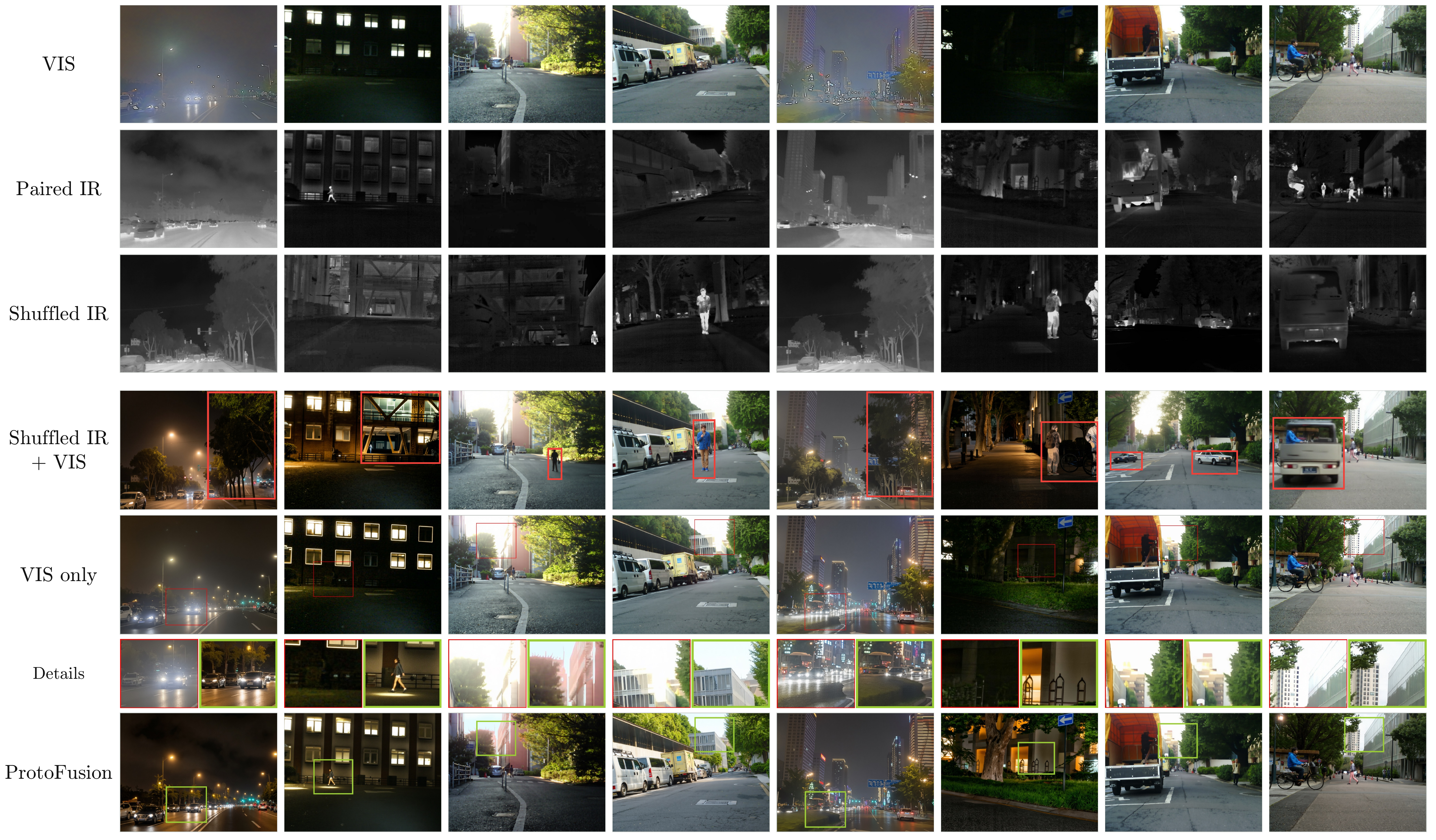}
    \vspace{0.4em}
    \caption{
Ablation study on the influence of infrared input. Each group shows the original IR/VIS inputs, a shuffled IR input, and the results generated under VIS-only, shuffled IR+VIS, and IR+VIS settings.
}
    \label{fig:onlyvis}
\end{figure*}

\subsection{Ablation Study}\label{sec:ablation}

\begin{table}[!t]
\centering
\caption{Protocol components on the FMB test pairs (UniPercept; higher is better; best in bold).}
\label{tab:pgpr_component_ablation}
\footnotesize
\setlength{\tabcolsep}{2pt}
\renewcommand{\arraystretch}{1.20}
\begin{tabular*}{\linewidth}{@{\extracolsep{\fill}}lrrrr@{}}
\toprule
Setting & IAA & IQA & ISTA & Overall\\
\midrule
Human Prompt & 29.1049 & 39.2397 & 46.0433 & 38.1293\\
Initial Protocol & 31.6094 & 40.7125 & 45.5683 & 39.2967\\
Principles Only & 30.6170 & 40.6623 & 45.9330 & 39.0708\\
w/o State Cards & 34.4786 & 44.3286 & 48.4580 & 42.4217\\
w/o Style Profile & 35.3647 & 45.3844 & 48.7804 & 43.1765\\
Full PGPR & \bestnum{39.8040} & \bestnum{52.3004} & \bestnum{51.2281} & \bestnum{47.7775}\\
\bottomrule
\end{tabular*}
\end{table}

\begin{table}[!t]
\centering
\caption{Refinement budget on all FMB test pairs (UniPercept; higher is better; best in bold).}
\label{tab:pgpr_rounds}
\footnotesize
\setlength{\tabcolsep}{2pt}
\renewcommand{\arraystretch}{1.20}
\begin{tabular*}{\linewidth}{@{\extracolsep{\fill}}lrrrr@{}}
\toprule
Budget & IAA & IQA & ISTA & Overall\\
\midrule
$T=1$ & 31.6094 & 40.7125 & 45.5683 & 39.2967\\
$T=3$ & 32.6509 & 44.7830 & 48.3902 & 41.9414\\
$T=6$ & {39.8040} & \bestnum{52.3004} & {51.2281} & \bestnum{47.7775}\\
$T=9$ & \bestnum{40.0701} & 51.5103 & \bestnum{51.2759} & 47.6188 \\
\bottomrule
\end{tabular*}
\end{table}

\noindent\textbf{Protocol Representation and Refinement.} Table~\ref{tab:pgpr_component_ablation} evaluates the contributions of protocol design and offline refinement using the same inputs, VLM, generator, random seeds, and generation settings. Overall improves from 38.13 with Human Prompt to 39.30 with Initial Protocol, and further reaches 47.78 with Full PGPR. Protocol initialization provides a modest gain of 1.17 points, whereas offline refinement contributes an additional 8.48 points. This substantial improvement highlights the importance of refining fusion instructions based on diagnosed failures, rather than relying on a fixed protocol. Notably, these gains are achieved without modifying the underlying generative models. Appendix~\ref{sec:supp_motivation_prompt_refinement} illustrates the fixed-prompt failure modes associated with this comparison.

The component ablations further demonstrate the importance of the protocol structure. Removing either state cards or the style profile reduces the Overall score by more than four points, with consistent declines in IAA, IQA, and ISTA. State cards help adapt fusion instructions to source conditions and modality reliability, while the style profile provides explicit control over the target appearance. The Principles Only setting also performs substantially worse than Full PGPR, indicating that grounding principles alone are insufficient. Together, these results highlight the complementary roles of source-conditioned guidance, grounding principles, and appearance control in determining what information to preserve and how to present it.

\vspace{0.5em}
\noindent\textbf{Refinement Budget.} 
As shown in Table~\ref{tab:pgpr_rounds}, increasing the offline refinement budget from $T=1$ to $T=6$ improves Overall from 39.30 to 47.78, demonstrating the benefit of iterative diagnosis and protocol refinement. Further increasing $T$ to 9 yields only marginal gains in some evaluation dimensions, while Overall slightly decreases. We therefore adopt $T=6$ to balance fusion performance and offline refinement cost. Since protocol optimization is performed offline, the inference procedure remains unchanged across different refinement budgets.

\begin{figure*}[t]
    \centering
    \includegraphics[width=\linewidth,trim=0 373 0 0,clip]{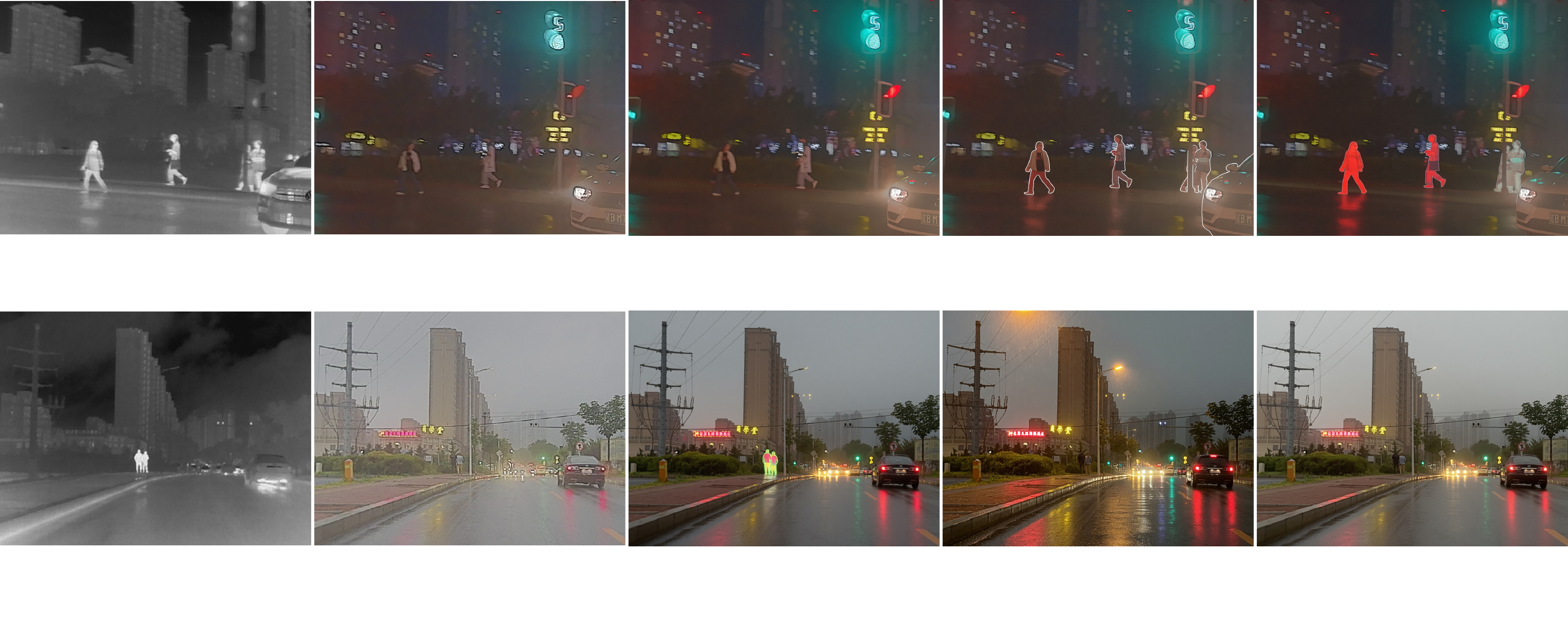}
    \par\smallskip
    {\fontsize{8}{9.2}\selectfont
    \setlength{\tabcolsep}{3pt}
    \begin{tabularx}{\linewidth}{*{5}{>{\centering\arraybackslash}X}}
    Infrared & Visible &
    ... Increase luminance/contrast of pedestrian regions for saliency while preserving visible realism. ... &
    ... Enhance luminance and contrast of \textcolor{red}{pedestrian shapes from IR} while preserving their VIS geometry ... &
    ... prioritize IR for defining sharp, crisp boundaries of the pedestrians. \textcolor{red}{Inject IR saliency into luminance and contrast}. \\
    \end{tabularx}}
    \par\medskip
    \includegraphics[width=\linewidth,trim=0 73 0 300,clip]{Fig05_PGPRv1.pdf}
    \par\smallskip
    {\fontsize{8}{9.2}\selectfont
    \setlength{\tabcolsep}{3pt}
    \begin{tabularx}{\linewidth}{*{5}{>{\centering\arraybackslash}X}}
    Infrared & Visible &
    ... \textcolor{red}{Overlay thermal saliency from the IR image to highlight pedestrians} and vehicles as bright, crisp shapes ... &
    ... Night urban street scene with \textcolor{red}{rain} ... &
    View Mode: enhanced reality. Preserve IR target shapes and boundaries; \textcolor{red}{avoid turning targets into blobs} ... \\
    \end{tabularx}}
    \caption{Visualization of the PGPR refinement trajectory. From left to right, we show the intermediate fused results across refinement rounds.}
    \label{fig:pgpr_iter}
\end{figure*}

\begin{table}[!t]
\centering
\caption{Prompt optimization and SBC comparison on the full FMB test set using UniPercept. Higher is better.}
\label{tab:ablation_sbc}
\footnotesize
\setlength{\tabcolsep}{2pt}
\renewcommand{\arraystretch}{1.15}
\begin{tabular*}{\linewidth}{@{\extracolsep{\fill}}lrrrr@{}}
\toprule
Method & IAA & IQA & ISTA & Overall\\
\midrule
Human Prompt & 29.1049 & 39.2397 & 46.0433 & 38.1293\\
\shortstack[l]{VLM-optimized Prompt} & 37.5263 & 38.2645 & 43.3942 & 39.7283\\
PGPR & \underline{39.8040} & \underline{52.3004} & \underline{51.2281} & \underline{47.7775}\\
PGPR+SBC & \bestnum{42.0143} & \bestnum{53.7696} & \bestnum{51.7683} & \bestnum{49.1841}\\
\bottomrule
\end{tabular*}
\end{table}

\par\smallskip\noindent\textbf{Protocol Optimization.} We further compare PGPR with open-ended VLM-based prompt optimization to examine the benefits of structured protocol refinement. Starting from the Human Prompt, the VLM-optimized Prompt baseline iteratively revises the prompt through unconstrained VLM-guided search, without predefined protocol slots or validated edit operations. As shown in Tab.~\ref{tab:ablation_sbc}, this baseline improves IAA but reduces IQA and ISTA, yielding only a modest gain in Overall. In contrast, PGPR improves all three dimensions and achieves an Overall score of 47.78, outperforming the VLM-optimized Prompt baseline by more than eight points. By refining structured fusion instructions through source-grounded diagnosis and validated edits, PGPR provides more consistent control over modality contributions, structural preservation, and visual appearance.

\par\smallskip\noindent\textbf{Role of the Infrared Input.} As shown in Fig.~\ref{fig:onlyvis}, we compare VIS-only, shuffled IR+VIS, and original IR+VIS generation under the same optimized protocol and generation settings. In the shuffled IR+VIS setting, we replace the original infrared input with an unrelated infrared image from another scene, while keeping the visible input unchanged. VIS-only generation can improve visual appearance but often misses infrared-salient targets and structures obscured in the visible input. In contrast, shuffled IR introduces substantial content alterations, including unrelated pedestrians, vehicles, and background structures that are absent from the original scene. Using the original IR input enables ProtoFusion to recover complementary infrared information while maintaining a coherent visible-light appearance. These comparisons demonstrate that infrared information directly influences the generated scene content, rather than merely serving as auxiliary guidance for visible image enhancement. The destructive effects of shuffled IR further highlight the importance of source-grounded guidance and cross-modal consistency control in generative image fusion.

\par\smallskip\noindent\textbf{Refinement Dynamics.}
To better understand how PGPR improves the fusion result, we visualize the intermediate outputs along the refinement trajectory in Fig.~\ref{fig:pgpr_iter}. The refinement trajectories illustrate how PGPR adjusts modality contributions in response to diagnosed generation errors.

In the first trajectory, the early outputs preserve the visible appearance but under-emphasize thermally salient pedestrians. Through protocol refinement, the prompt progressively strengthens the use of infrared evidence for pedestrian regions, leading to clearer target responses and more distinguishable boundaries. In the second trajectory, directly injecting infrared saliency improves target visibility but may also introduce over-enhanced or blob-like structures. PGPR then introduces more constrained instructions, encouraging the model to preserve infrared target shapes while avoiding unnatural target appearance. Together, these examples show how PGPR adjusts both target emphasis and rendering constraints through iterative refinement.

\begin{figure}[!t]
    \centering
    \begin{minipage}{0.86\linewidth}
    \centering
    \includegraphics[width=\linewidth,trim=0 0 1209.5bp 0,clip]{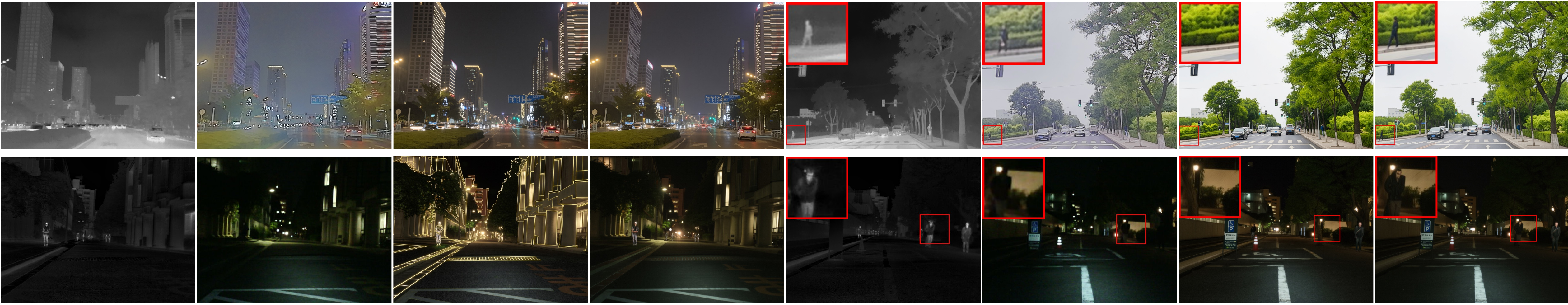}
    \par\smallskip
    \includegraphics[width=\linewidth,trim=1209.5bp 0 0 0,clip]{Fig06_SBC.pdf}
    \par\smallskip
    {\fontsize{8}{9.2}\selectfont
    \setlength{\tabcolsep}{0pt}
    \begin{tabularx}{\linewidth}{*{4}{>{\centering\arraybackslash}X}}
    Infrared & Visible & PGPR & \textbf{PGPR+SBC}\\
    \end{tabularx}}
    \end{minipage}
    \caption{Visual comparison of PGPR and PGPR+SBC on representative cases. SBC adjusts local target emphasis and appearance through a single prompt correction.
    }
    \label{fig:sbc_visual}
\end{figure}

\Needspace{7\baselineskip}
\par\smallskip\noindent\textbf{SBC Correction.}
SBC further complements PGPR by addressing image-specific errors that remain after protocol optimization. It analyzes region-level modality reliability and cross-modal conflicts to guide conservative prompt corrections. As shown in Fig.~\ref{fig:sbc_visual}, SBC adjusts target emphasis and appearance in representative cases, further improving Overall to 49.18 in Tab.~\ref{tab:ablation_sbc}. The PGPR baseline matches Full PGPR in Tab.~\ref{tab:pgpr_component_ablation} and $T=6$ in Tab.~\ref{tab:pgpr_rounds}, with all three studies evaluated on the FMB test set. These results highlight the complementary benefits of structured protocol optimization and targeted correction of residual errors.
\par

\begin{figure}[!t]
    \centering
    \captionsetup[subfigure]{justification=centering,singlelinecheck=false,skip=4pt}
    \begin{subfigure}[b]{\linewidth}
        \centering
        \includegraphics[width=\linewidth]{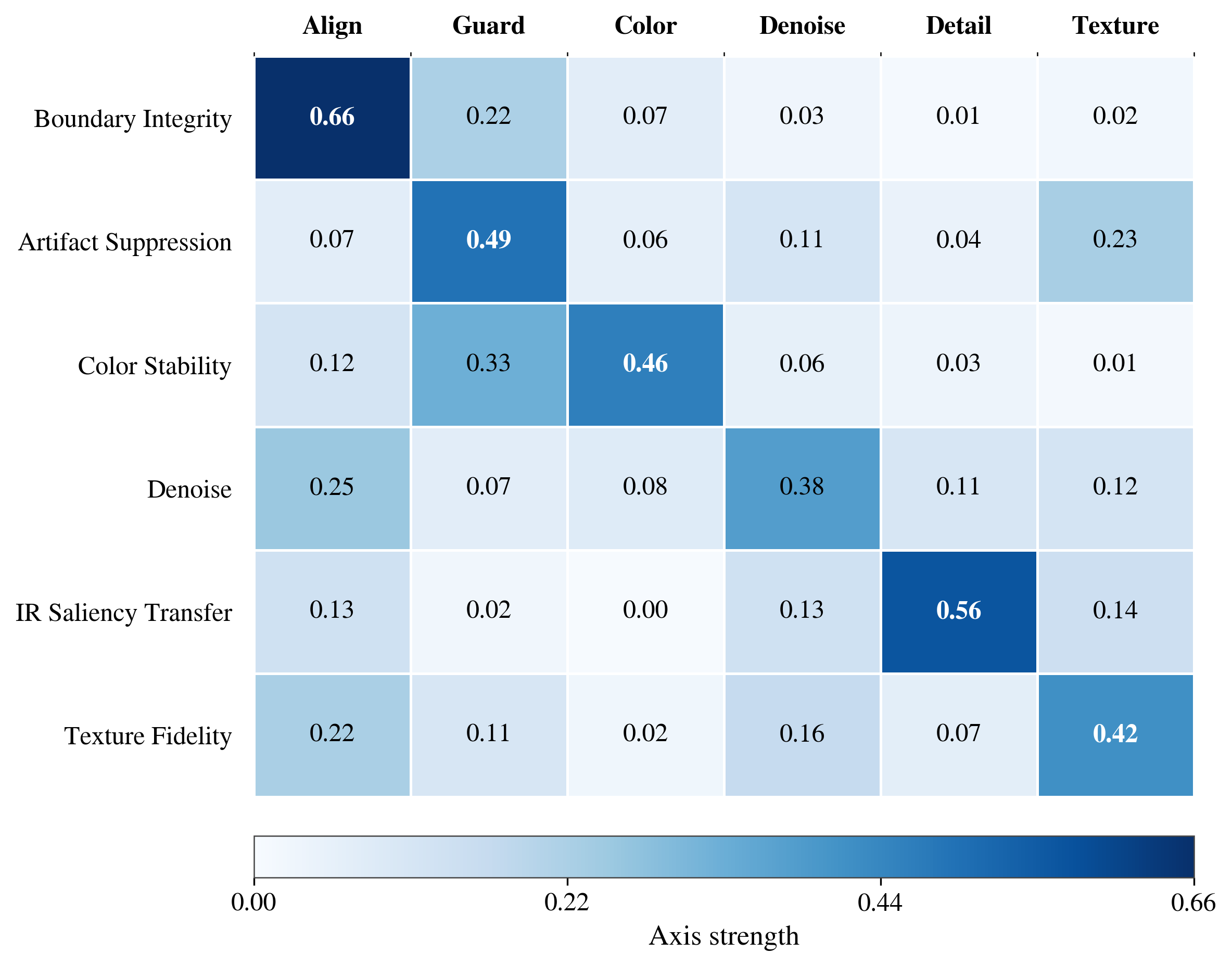}
        \caption{Intrinsic functional factors}
        \label{fig:intrinsic_factor_map}
    \end{subfigure}
    \par\medskip
    \begin{subfigure}[b]{\linewidth}
        \centering
        \includegraphics[width=\linewidth]{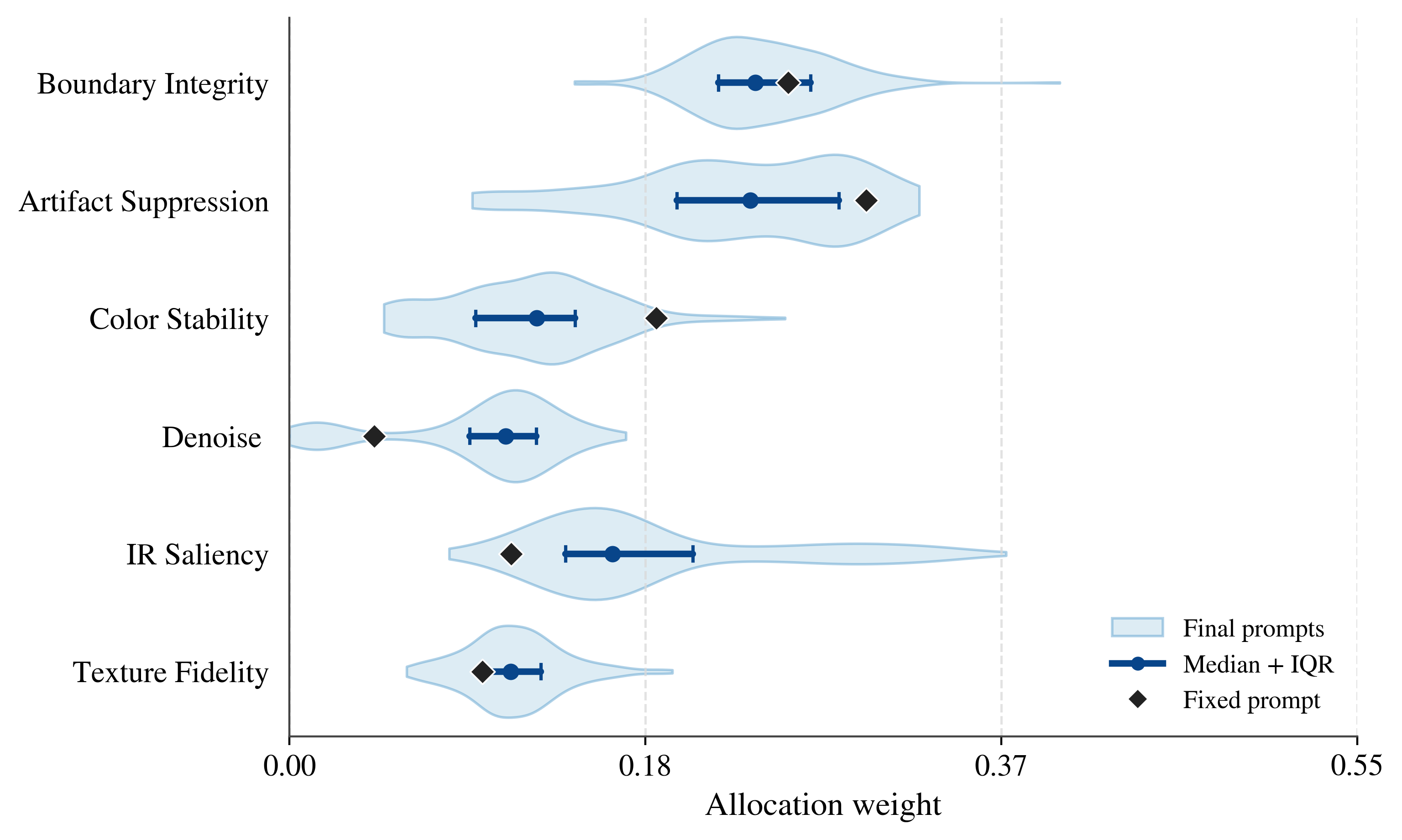}
        \caption{Prompt-level control allocation}
        \label{fig:control_allocation_profile}
    \end{subfigure}
    \caption{Analysis of the learned protocol. (a) Intrinsic functional factors. (b) Prompt-level control allocation. Blue distributions show the final prompts; blue points and bars show the median and interquartile range; black diamonds denote the human-crafted prompt.}
    \label{fig:protocol_analysis_main}
\end{figure}

\par\smallskip\noindent\textbf{Protocol Analysis.}
We analyze the learned protocol in terms of both its intrinsic functional structure and its prompt-level instantiation. The protocol is decomposed into intermediate functional components, projected into a shared functional signature space, and clustered into intrinsic factors. As shown in Fig.~\ref{fig:intrinsic_factor_map}, the most informative pattern is the \ensuremath{<}IR Saliency Transfer\ensuremath{>} factor, which is concentrated on the detail axis and largely decoupled from color-related behavior. This suggests that the learned protocol assigns IR information a selective role, contributing salient structural details while leaving visible appearance regulation to color-related controls.

We further parse the final prompts for the FMB dataset into semantic control units and normalize them over the same functional axes. Figure~\ref{fig:control_allocation_profile} shows that protocol-derived prompts occupy a bounded allocation range around the human-crafted prompt, with variations concentrated on specific control axes. Together, these results indicate that the learned protocol organizes fusion intent into separable functional factors and instantiates them as controlled prompt-level reweighting before generation.

\subsection{Cross-generator Protocol Transfer}\label{sec:generator_transfer}

The main quantitative experiments use Qwen-Image-Edit as the image generator. To examine whether the learned protocol can be reused with another generator, we replace Qwen-Image-Edit with GPT-Image-2 while retaining the same optimized protocol. We directly reuse the protocol optimized with Qwen-Image-Edit for GPT-Image-2. This comparison therefore examines cross-generator reuse of a fixed learned protocol within the ProtoFusion formulation.

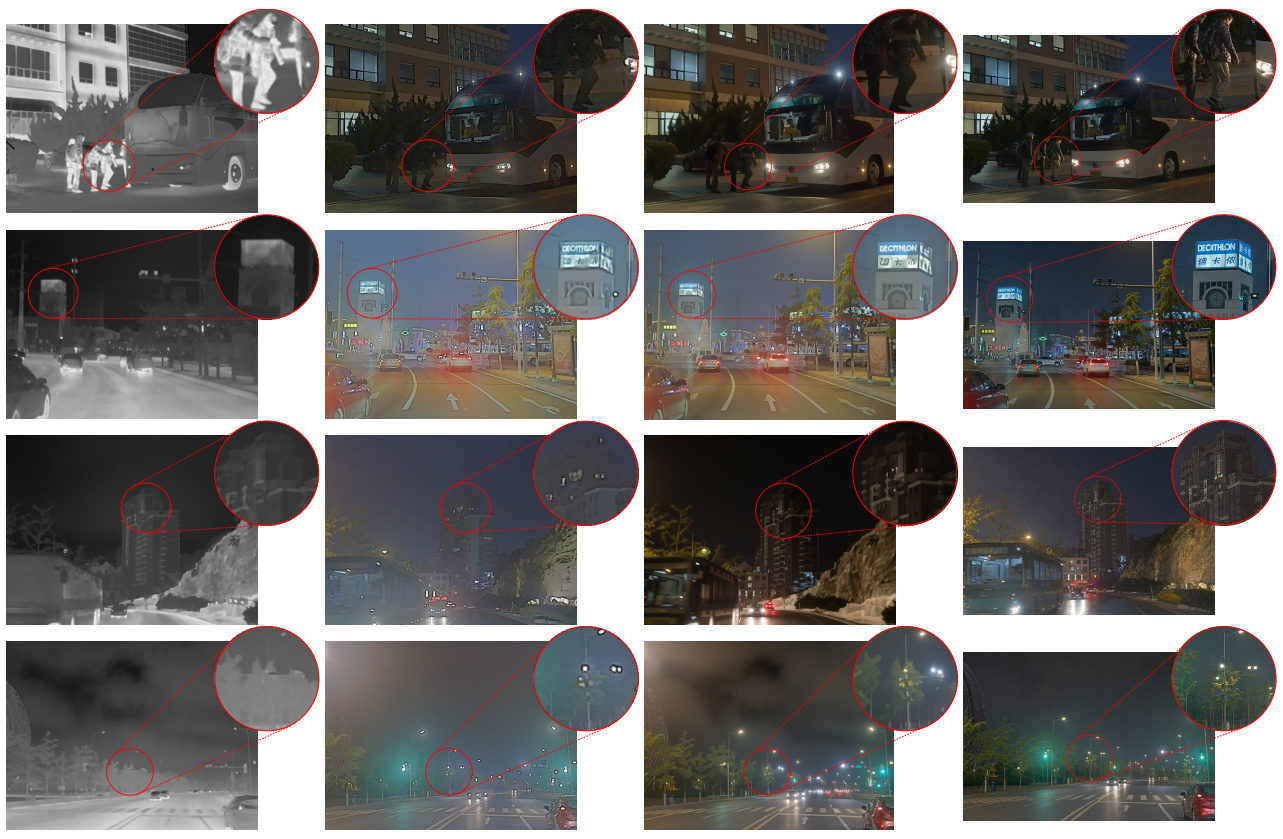
\begin{figure}[t]
    \centering
    \begin{tikzpicture}
        \node[anchor=south west, inner sep=0] (img) at (0,0)
        {\includegraphics[width=\linewidth]{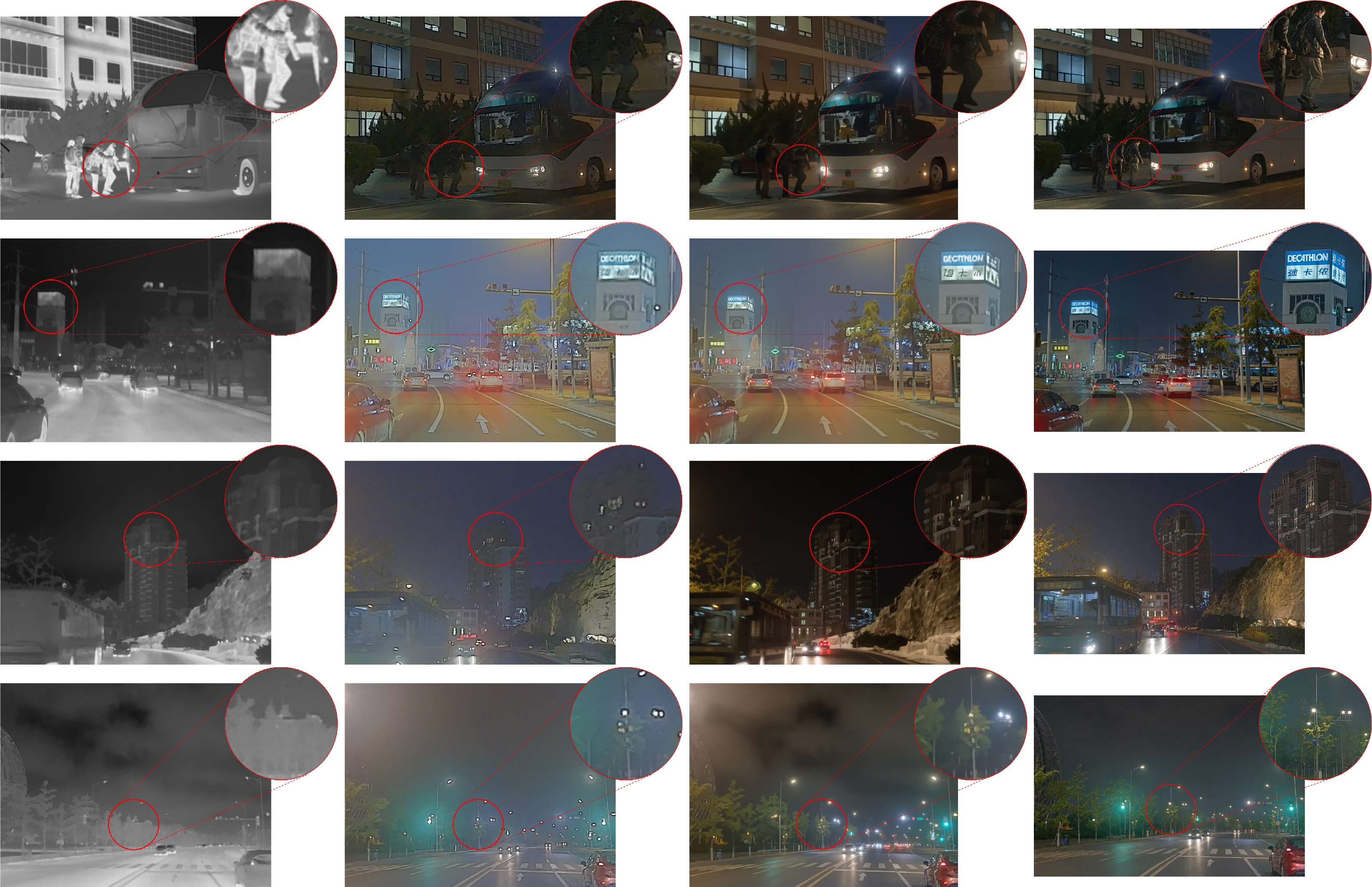}};
        \begin{scope}[x={(img.south east)}, y={(img.north west)}]
            \node[anchor=north, align=center, font=\fontsize{8}{9.2}\selectfont] at (0.10,0.0) {Infrared};
            \node[anchor=north, align=center, font=\fontsize{8}{9.2}\selectfont] at (0.35,0.0) {Visible};
            \node[anchor=north, align=center, font=\fontsize{8}{9.2}\selectfont] at (0.60,0.0) {ProtoFusion\\(Qwen)};
            \node[anchor=north, align=center, font=\fontsize{8}{9.2}\selectfont] at (0.85,0.0) {ProtoFusion\\(GPT)};
        \end{scope}
    \end{tikzpicture}
    \vspace{-2em}
    \caption{Comparison of image generators under a fixed learned protocol. The columns show the infrared and visible inputs, followed by ProtoFusion with Qwen-Image-Edit and GPT-Image-2. The protocol learned with Qwen-Image-Edit is reused without reoptimization.}
    \label{fig:gptimage2}
\end{figure}

Figure~\ref{fig:gptimage2} compares the two instantiations on the same infrared and visible image pairs. Both produce visible-style outputs that emphasize infrared-supported targets, while the underlying generator affects their appearance. In these nighttime examples, the GPT-Image-2 outputs show cleaner local structures and sharper rendering around pedestrians, building contours, roadside signs, and street-light regions. The differences indicate that protocol guidance and generator capability jointly influence the final result. 

\suppressfloats[t]
\begin{table}[t]
\centering
\caption{Inference latency per image pair, including prompt composition and one image-generation step (excluding optional SBC).}
\label{tab:runtime}
\small
\setlength{\tabcolsep}{3pt}
\renewcommand{\arraystretch}{1.15}
\begin{tabular*}{\linewidth}{@{\extracolsep{\fill}}lr@{}}
\toprule
Generator configuration & Time (s/pair)\\
\midrule
Qwen-Image-Edit & $\sim$188\\
Qwen-Image-2.1-Turbo & $\sim$9\\
\bottomrule
\end{tabular*}
\end{table}

These examples support a modular interpretation of ProtoFusion: the learned protocol specifies \emph{what} evidence should be preserved, emphasized, or suppressed, while the generator determines \emph{how} these instructions are rendered. Reusing the protocol without reoptimization demonstrates its applicability to a second generator in the illustrated cases. 
These examples demonstrate that the optimized protocol can guide different image generators. A matched quantitative evaluation would further assess cross-generator performance in perceptual quality, source consistency, and downstream utility.

\subsection{Inference Runtime}\label{sec:efficiency}
ProtoFusion optimizes a shared fusion protocol offline and reuses it across test images, avoiding repeated protocol refinement during inference. With the optimized protocol fixed, inference requires only prompt composition and image generation.

As shown in Tab.~\ref{tab:runtime}, the Qwen-Image-Edit implementation takes approximately 188\,s per image pair, including 38\,s for prompt composition and 150\,s for image generation. We further evaluate Qwen-Image-2.1-Turbo~\citep{qwen2026image21turbo}, a newer and more lightweight generative model with a 7B-parameter image generation backbone and only eight denoising steps. Under a separate execution configuration, prompt composition and image generation together take approximately 9\,s per pair. These results show that image generation dominates the computational cost in the original implementation, while the optimized protocol can be reused with more efficient generative models.

\vspace{-0.1em}
\subsection{Limitations and Discussion}\label{sec:limitations}

\noindent\textbf{Hallucination and fidelity.} ProtoFusion aims to generate fused images with a natural visible appearance for practical visual interpretation. Under severe visible degradation, infrared inputs can reveal salient structures but cannot fully determine their visible appearance. The generator may therefore introduce visually plausible details that are unsupported by either source. Although PGPR and SBC reduce source inconsistencies, residual errors persist (Table~\ref{tab:fmb-four-category-errors}). Recent advances in image editing, such as GPT-Image-2.5, have improved reference-image preservation and editing precision \citep{openai2026images25}. As visual grounding and instruction following improve, such hallucinations may become less frequent without changing the core PGPR formulation.

\par\smallskip\noindent\textbf{Protocol feedback and control.} PGPR relies on VLM diagnoses and proxy scores that may overlook source inconsistencies. Its best-so-far selection ensures a non-decreasing retained proxy score, but cannot guarantee improved source fidelity. SBC further identifies residual errors through region-level analysis, yet implements corrections through language instructions rather than direct image constraints. Consequently, a narrowly scoped prompt edit may alter image regions beyond its intended scope. Stronger image-level constraints and verification could make these corrections more reliable.

\par\smallskip\noindent\textbf{Computational cost.} The current Qwen-Image-Edit implementation requires approximately 188\,s per image pair for prompt composition and generation. With the newer Qwen-Image-2.1-Turbo, the observed base-path latency drops to approximately 9\,s (Section~\ref{sec:efficiency}). Although these measurements exclude the optional SBC module and real-time fusion remains challenging, this substantial improvement highlights the potential of more efficient generative backbones. Continued advances in generative foundation models are expected to further improve the inference efficiency and practical applicability of ProtoFusion.

\FloatBarrier
\section{Conclusion}\label{sec:conclusion}

We presented ProtoFusion, a protocol-guided generative framework for infrared and visible image fusion. ProtoFusion replaces fixed fusion operator design with protocol-constrained generation and uses offline refinement to optimize a structured fusion protocol that is fixed at inference time. The optional SBC module further addresses residual source inconsistencies through targeted corrections. Experiments on standard benchmarks show that this design yields strong quantitative performance, higher UniPercept scores, and better visual quality, particularly when the visible modality is degraded.

More broadly, ProtoFusion introduces a new paradigm for image fusion. It makes fusion intent explicit and optimizes a structured protocol to guide foundation-model generation. This shifts the focus from designing fixed fusion operators to controlling what information to preserve and how to present it. Source fidelity and computational cost remain challenges. However, continued advances in foundation models may further improve fusion quality, reliability, and efficiency. ProtoFusion can benefit from these advances without changing its core framework. We hope this paradigm inspires further research on foundation-model-driven image fusion.

\FloatBarrier

\section*{Data Availability}
The FMB, MSRS, and RoadScene datasets used in this study are publicly available.

\section*{Conflict of Interest}
The authors affirm that there are no commercial or associative relationships that could be perceived as a conflict of interest related to the submitted work.

\bibliographystyle{spbasic}
\AddToHookNext{shipout/after}{\balance}
\bibliography{sn-bibliography}

\clearpage
\nobalance
\appendix
\captionsetup[figure]{skip=6pt}
\setlength{\textfloatsep}{10pt plus 2pt minus 2pt}
\setlength{\dbltextfloatsep}{12pt plus 2pt minus 2pt}
\renewcommand{\thesection}{\Alph{section}}
\section*{Appendix}
\suppressfloats[t]
\setcounter{dbltopnumber}{2}
\renewcommand{\dbltopfraction}{0.95}
\renewcommand{\topfraction}{0.9}
\renewcommand{\textfraction}{0.1}
Appendix~\ref{app:additional} gives the implementation details and prompt templates. Appendix~\ref{app:additional_results} supplements the main evaluation with source-referenced fidelity metrics and additional nighttime comparisons. Appendix~\ref{app:failure_analysis} examines fixed-prompt failures and hallucination propagation. The component, refinement-budget, and SBC ablations use the FMB test pairs; the remaining benchmark comparisons use the test-set sizes stated in Section~\ref{sec:setup}.

\section{Implementation and Prompt Details}\label{app:additional}
\lstset{basicstyle=\footnotesize\ttfamily,breaklines=true,lineskip=-0.2pt,
  breakatwhitespace=true,columns=fullflexible,keepspaces=true,
  showstringspaces=false}

\subsection{Region-level Evidence for SBC}
\label{sec:supp_sbc_details}

\paragraph{Structured Belief Construction.}
In our implementation, SBC constructs the belief state in Eq.~\ref{eq:sbc_state} through a hybrid region-level diagnosis. We first use SAM to obtain candidate regions and treat them only as local analysis units rather than fusion masks. For each region, we extract aligned crops from the visible image, infrared image, and the selected PGPR output. SBC then computes traditional visual cues to provide grounded evidence, including sharpness, contrast, brightness balance, cleanliness, edge density, thermal dynamic range, hotspot fraction, and hotspot compactness. These measurements are used to describe whether the visible modality provides reliable texture and appearance, whether the infrared modality provides compact thermal targets and clear boundaries, and whether the PGPR output preserves or distorts the source evidence. In parallel, Qwen3-VL-8B-Instruct parses the same region crops into structured semantic evidence, including presence, location, appearance, shape, and consistency cues, together with reliability, conflict-risk, and confidence estimates. The final belief combines measurable regional cues with semantic interpretation from Qwen3-VL-8B-Instruct.

After obtaining the region-level belief state, SBC maps it to the conservative prompt correction defined in Eq.~\ref{eq:sbc_edit}. The belief state is summarized into image-level statistics, such as modality reliability, average confidence, average conflict, infrared-supported regions, high-confidence regions, and ambiguous regions. 
These statistics determine a conservative correction policy, including how strongly infrared evidence should be exposed, how strictly boundaries should be preserved, and whether the global appearance should follow the visible image, the PGPR prior, or a balanced rendering strategy. A small number of representative regions are further selected as local anchors, such as infrared-supported salient targets, stable structural boundaries, texture-supported visible regions, or ambiguous cross-modal regions. Finally, Qwen3-VL-8B-Instruct compiles these global policies and local anchor hints into a revised prompt, while treating the selected PGPR prompt as the authoritative base prompt. Thus, SBC only performs a scoped correction after PGPR: it may strengthen missing thermal targets, suppress thermal blobs, protect boundaries, or avoid visible degradation leakage, but SBC preserves the global fusion intent and applies a scoped correction to the selected prompt.

\subsection{Diagnostic and Update Rules}\label{sec:diagnostic_rules}
The diagnostic score combines semantic fidelity, artifact control, texture, information gain, and protocol compliance with weights 0.30, 0.25, 0.20, 0.15, and 0.10, respectively. Nine source-grounded checks return binary judgments and confidence estimates. Detected fatal source violations receive a score of $-1$ and are excluded from inheritance; a strict pass requires passing all nine checks and achieving a score of at least 0.80.

Starting from an evaluated non-fatal incumbent, the editor proposes one allowed slot edit: add an environment-state item, revise a principle, or add a principle. The style profile stays fixed. The validator rejects malformed, empty, repeated, or ineffective edits, as well as edits with invalid conditions. A valid proposal is composed, generated, and diagnosed before selection. Lower-scoring or fatal candidates do not replace the incumbent; later edits start from a best-scoring valid state. The acceptance rule tracks the retained critic score, while source fidelity is evaluated separately.

\subsection{Protocol and Prompt Details}
\label{sec:supp_protocol_prompt}

This section presents the fixed prompt baseline and the initial protocol used by ProtoFusion. 
The fixed human-crafted prompt is used in the failure examples in Fig.~\ref{fig:fixed_prompt_failure} and the controlled prompt/protocol comparison in Table~\ref{tab:pgpr_component_ablation}. 
It is directly fed to the generative fusion model and kept unchanged for all image pairs. 
In contrast, ProtoFusion starts from an initial structured protocol, which specifies the environment variables, fusion principles, and style constraints for prompt composition. 
This protocol is subsequently optimized by PGPR via iterative evaluation-guided protocol refinement, and the optimized protocol is fixed for inference.

\noindent\textbf{Fixed Human-crafted Prompt.}
The fixed prompt is used to evaluate one-shot prompting without protocol refinement. The human-crafted prompt applies the same fusion instructions to every image pair, whereas PGPR combines image-specific prompt composition with an optimized fusion protocol.

{\small
\begin{lstlisting}
You are doing image fusion.

Inputs:
- Image 1: Color visible (VIS)
- Image 2: Grayscale infrared (IR)

Goal:
Generate ONE fused image.

Rules:
1. Use the VIS image to define scene structure, layout, edges, 
   and background whenever possible.
2. If the VIS image is overexposed, underexposed, or low-contrast, 
   use IR luminance to add texture, 
   but render these details with a neutral visible appearance.
3. No thermal colors (red, orange, or yellow). No glow. No halo.
4. Do not let the IR image define global structure or background.
5. Keep VIS colors, textures, and fine details as the main appearance.
6. The result should look like a normal visible-light photograph.
7. No IR-style background, no double edges, no noise, and no ghosting.
\end{lstlisting}
}

\noindent\textbf{Initial Protocol for PGPR.}
PGPR starts from the following structured protocol. 
In contrast to the fixed prompt, the protocol is not itself the final generation instruction. It first extracts an environment state from the input pair, applies condition-dependent fusion principles, defines the target visual style, and then assembles an operational prompt for the generator. 
During refinement, PGPR updates the protocol slots and principle descriptions according to multimodal evaluation feedback, so that later prompts are composed from an optimized protocol rather than from a fixed manually written instruction.

\Needspace{12\baselineskip}
{\small
\begin{lstlisting}
Role: You are the Fusion Logic Engine.
Input: Image 1 (Visible/Vis), Image 2 (Infrared/IR).
Goal: Analyze the images and generate a specific prompt for a Text/Image-to-Image 
model (Qwen-Image) to create a fused image that preserves:
(1) Vis color/texture realism,
(2) clean, single boundaries (no ghosting/bleed/smear), and
(3) information gain (saliency vs Vis, detail vs IR).

Phase 1: Perception Layer (Variable Extraction)
Observe visible and infrared images and fill in the [Environment State Card].
Do not describe; just extract values.

[Environment State Card]
1. {Lighting_Condition}: [Day / Dusk / Evening / Night / Artificial_Light]
   - Condition: only rely on Vis image to judge.
2. {Vis_Quality}: [High_Fidelity / Overexposed / Low_Light_Noise]
3. {IR_Targets}: [None / Pedestrians / Vehicles / Machinery]
4. {Information_Dominance}: [Balanced / Vis_Lead / IR_Lead]
5. {Vis_Edge_Reliability}: [Reliable / Unreliable]
   - Reliable: edges are crisp and consistent (not dominated by noise/overexposure).
6. {Chroma_Bleed_Risk}: [Low / High]
   - High: strong colored light sources (neon, brake lights, traffic lights)
     likely to bleed.
7. {Alignment_Risk}: [Low / High]
   - High: IR and Vis structures appear slightly shifted or mismatched around
     object boundaries.

Phase 2: Decision Layer (Principle Application)
Based on Phase 1 variables, apply the following Fusion Principles.
Write the final prompt as a short, operational plan:
clear steps + a compact list of constraints.

Principle A (Night Robustness and Noise Control):
IF {Lighting_Condition} is Night AND {Vis_Quality} is Low_Light_Noise:
- Strategy:
  1) Use Vis as the base for overall scene appearance (global tone and colors),
     but treat very dark/noisy Vis regions as low-trust for fine texture.
  2) Preserve IR target shapes and boundaries; avoid turning targets into blobs.
  3) Apply edge-preserving denoise (remove grain without smearing edges or textures).
- Constraint:
  - No painterly smoothing or texture smear.
  - No synthetic grain; do not amplify noise.
  - Preserve crisp edges; do not blur boundaries.

Principle B (Thermal Saliency with Color Discipline):
IF {IR_Targets} exist:
- Strategy:
  1) Make IR targets easier to spot by increasing luminance/contrast of the 
     target regions (saliency), while keeping Vis realism.
  2) Preserve target geometry (shape and boundary consistency) without double
     contours.
  3) If {Chroma_Bleed_Risk} is High: lock chroma to Vis; inject IR only into
     luminance/contrast (avoid colored thermal halos).
- Constraint:
  - No hallucinated objects or structures absent in both IR and Vis.
  - No colored halos around targets; no color leakage across boundaries when
    {Chroma_Bleed_Risk} is High.

Principle C (Texture Preservation without Smear):
IF {Vis_Quality} is High_Fidelity:
- Strategy:
  1) Preserve Vis micro-textures (text, lines, road texture) and natural color
     statistics.
  2) Add IR saliency without washing out high-frequency Vis details.
  3) Prefer sharp detail preservation over global smoothing.
- Constraint:
  - No over-smoothing; no oil-painting look.
  - Do not wash out fine Vis details when adding IR saliency.

Principle D (Boundary Alignment Policy):
IF {IR_Targets} exist OR {Alignment_Risk} is High:
- Strategy:
  1) Enforce a single contour per object (avoid double edges / semi-transparent 
     outlines).
  2) If {Vis_Edge_Reliability} is Reliable AND {Alignment_Risk} is Low:
     anchor object boundaries to Vis edges (hard anchor).
     Else: use a soft alignment; prioritize one boundary source
     (prefer IR boundary for targets) and suppress secondary contours.
  3) Prevent offset blending of contours; keep boundaries geometrically consistent.
- Constraint:
  - No ghost edges, double contours, or transparent overlays.
  - No offset blending that creates duplicated boundaries.

Phase 3: Stylization Layer (Visual Definition)
Define the artistic parameters:
- View Mode: Realistic Night Vision / Enhanced Reality
- Color Palette: Natural Vis colors
- Clarity: 4K Resolution, Sharp Edges

Phase 4: Construction Layer (Prompt Assembly)
Output the Final Prompt based on the previous layers.

[Final Prompt]: <Write the English prompt here>
\end{lstlisting}
}

\twocolumn[{
\section{Additional Experimental Results}\label{app:additional_results}
\begin{minipage}{\textwidth}
\captionsetup{hypcap=false}
\centering
\captionof{table}{Additional source-referenced evaluation. Lower CE and higher $Q_{AB/F}$ and VIF are better. Best and second-best results are bold and underlined, respectively.}
\label{tab:traditional_metrics}
\footnotesize
\renewcommand{\arraystretch}{1.12}
\setlength{\tabcolsep}{4pt}
\begin{tabular*}{\linewidth}{@{\extracolsep{\fill}}llrrrrrrrrr@{}}
\toprule
& & \multicolumn{3}{c}{FMB} & \multicolumn{3}{c}{MSRS} & \multicolumn{3}{c}{RoadScene}\\
\cmidrule(lr){3-5}\cmidrule(lr){6-8}\cmidrule(lr){9-11}
Method & Venue & CE$\downarrow$ & $Q_{AB/F}\uparrow$ & VIF$\uparrow$ & CE$\downarrow$ & $Q_{AB/F}\uparrow$ & VIF$\uparrow$ & CE$\downarrow$ & $Q_{AB/F}\uparrow$ & VIF$\uparrow$\\
\midrule
CDDFuse & CVPR'23 & 1.8371 & 0.3500 & 0.2150 & \underline{1.0234} & 0.3721 & \underline{0.3171} & 0.9237 & 0.2460 & 0.1810 \\
DDFM & ICCV'23 & 1.8333 & 0.1131 & 0.0437 & 1.2518 & 0.2632 & 0.2482 & 1.4241 & 0.1282 & 0.0583 \\
TIMFusion & TPAMI'24 & 2.1071 & 0.2698 & 0.1481 & 1.7597 & 0.2552 & 0.2019 & \underline{0.6055} & 0.1978 & 0.1559 \\
Text-IF & CVPR'24 & \underline{1.7777} & \underline{0.3621} & \bestnum{0.2259} & 1.0391 & \underline{0.3778} & \bestnum{0.3269} & 0.8612 & \underline{0.2759} & \underline{0.1967} \\
C2RF & IJCV'25 & 1.8749 & 0.1847 & 0.1158 & 1.2196 & 0.1587 & 0.1348 & 1.0004 & 0.2652 & 0.1693 \\
DCEvo & CVPR'25 & 1.9175 & 0.3527 & 0.2153 & 1.0867 & 0.3769 & 0.3128 & 1.0545 & 0.2683 & 0.1792 \\
Deno-IF & NeurIPS'25 & 2.1191 & 0.2758 & 0.1747 & 1.3868 & 0.3098 & 0.2848 & 0.9421 & 0.2484 & 0.1851 \\
LUT-Fuse & ICCV'25 & 2.0738 & 0.2973 & 0.1886 & 1.2540 & 0.3380 & 0.2864 & 1.3454 & 0.2194 & 0.1610 \\
APTP & CVPR'26 & 1.9288 & 0.3345 & 0.1993 & \bestnum{1.0084} & 0.3437 & 0.2870 & 1.0902 & 0.2402 & 0.1867 \\
DPOFusion & CVPR'26 & 2.1022 & 0.2371 & 0.1664 & 3.3510 & 0.1931 & 0.2612 & 0.9711 & 0.2065 & 0.1679 \\
DRFusion & CVPR'26 & 1.8506 & 0.2595 & 0.1556 & 1.4457 & 0.2575 & 0.1974 & \bestnum{0.4930} & 0.2062 & 0.1473 \\
ISFuse & CVPR'26 & 2.3855 & 0.3102 & 0.1802 & 1.3161 & 0.3257 & 0.2618 & 1.7310 & 0.2368 & 0.1509 \\
\textbf{ProtoFusion } & & \bestnum{1.2906} & \bestnum{0.3641} & \underline{0.2253} & 1.1521 & \bestnum{0.3943} & 0.3052 & 0.9346 & \bestnum{0.3220} & \bestnum{0.2908} \\
\bottomrule

\end{tabular*}
\end{minipage}
\par\vspace{12pt}
\begin{minipage}{\textwidth}
\captionsetup{hypcap=false}
\centering
\includegraphics[width=0.493\linewidth,trim=0 1050.0bp 0 0.0bp,clip]{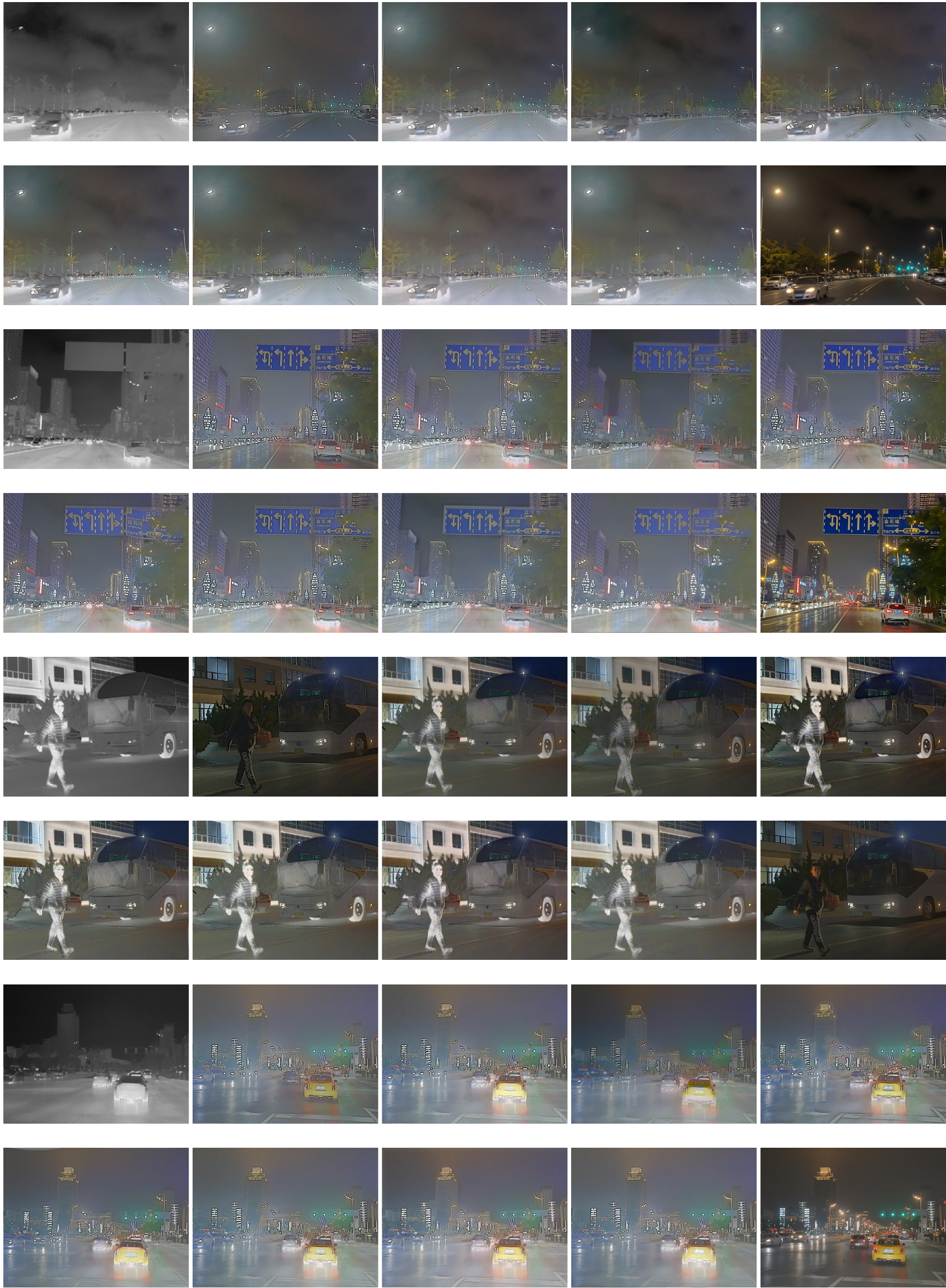}\hfill
\includegraphics[width=0.493\linewidth,trim=0 899.8bp 0 150.5bp,clip]{Fig10_final_visual.pdf}\par\smallskip
\includegraphics[width=0.493\linewidth,trim=0 749.6bp 0 300.8bp,clip]{Fig10_final_visual.pdf}\hfill
\includegraphics[width=0.493\linewidth,trim=0 599.3bp 0 451.1bp,clip]{Fig10_final_visual.pdf}\par\smallskip
\includegraphics[width=0.493\linewidth,trim=0 449.7bp 0 601.3bp,clip]{Fig10_final_visual.pdf}\hfill
\includegraphics[width=0.493\linewidth,trim=0 299.5bp 0 751.6bp,clip]{Fig10_final_visual.pdf}\par\smallskip
\includegraphics[width=0.493\linewidth,trim=0 149.2bp 0 901.8bp,clip]{Fig10_final_visual.pdf}\hfill
\includegraphics[width=0.493\linewidth,trim=0 0.0bp 0 1052.1bp,clip]{Fig10_final_visual.pdf}
\par\smallskip
    \fontsize{8}{9.2}\selectfont
    \setlength{\tabcolsep}{0pt}
    \renewcommand{\arraystretch}{0.95}
    \begin{tabularx}{0.99\linewidth}{
    *{10}{>{\centering\arraybackslash}X}
    }
    \oneline{Infrared}
    & \oneline{Visible}
    & \twoline{CDDFuse}{\confshort{CVPR}{23}}
    & \twoline{DDFM}{\confshort{ICCV}{23}}
    & \twoline{Text-IF}{\confshort{CVPR}{24}}
    & \twoline{C2RF}{\confshort{IJCV}{25}}
    & \twoline{DCEvo}{\confshort{CVPR}{25}}
    & \twoline{Deno-IF}{\confshort{NeurIPS}{25}}
    & \twoline{LUT-Fuse}{\confshort{ICCV}{25}}
    & \twoline{\textbf{Proto}}{\textbf{Fusion}} \\
    \end{tabularx}

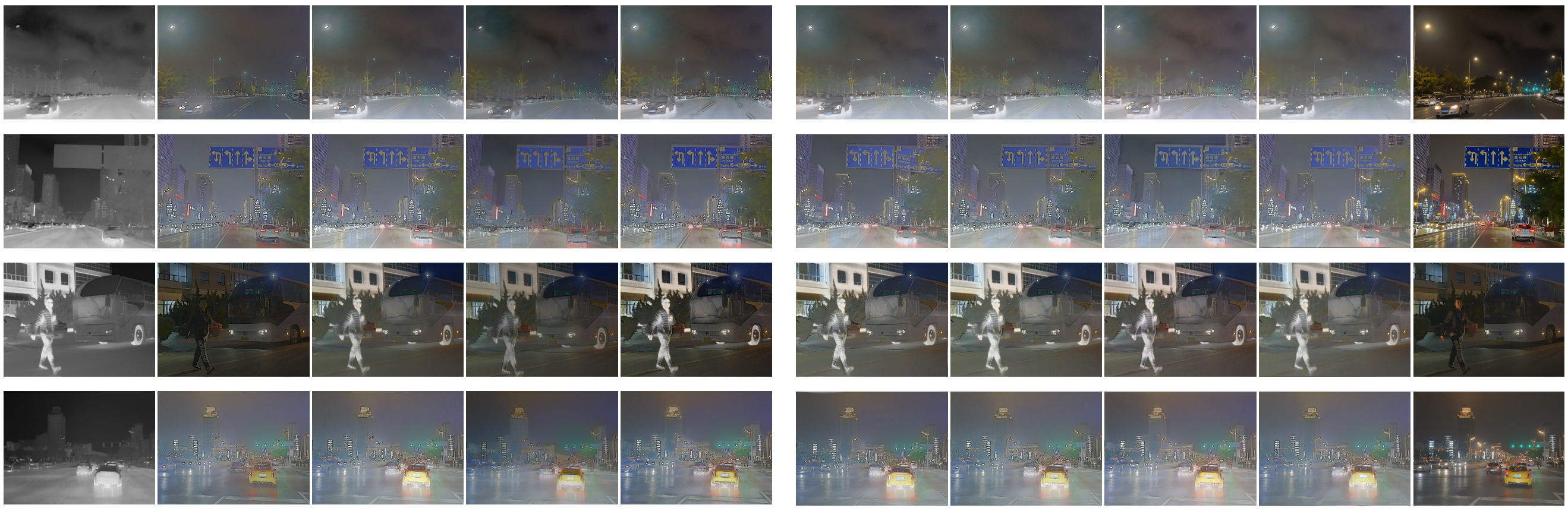
\captionof{figure}{Qualitative comparisons on degraded nighttime scenes.}
\label{fig:final_visual}
\end{minipage}
\vspace{12pt}
}]

\subsection{Traditional Fusion Metrics}\label{sec:source_fidelity} 
Traditional fusion metrics mainly assess how much low-level information a fused image preserves from its infrared and visible sources. This evaluation only partly matches the objective of ProtoFusion. Our framework deliberately generates images with a natural visible-light appearance. It uses infrared information to guide the rendering of salient targets and structures, instead of directly reproducing infrared pixel intensities. Consequently, metrics that favor low-level agreement with both sources may not fully capture the information conveyed by the generated images. We report CE, $Q_{AB/F}$, and VIF on three benchmarks to evaluate source discrepancy, edge transfer, and visual information fidelity, respectively. 

Despite this difference, ProtoFusion achieves the highest $Q_{AB/F}$ on all three datasets, the lowest CE on FMB, and the highest VIF on RoadScene, as shown in Tab.~\ref{tab:traditional_metrics}. It also achieves the second-highest VIF on FMB. The consistent advantage in $Q_{AB/F}$ is particularly notable, showing that our visible-light rendering can still preserve important edge information from both modalities. Although CE and VIF vary across datasets, these results demonstrate that ProtoFusion can retain substantial source information while pursuing a different visual representation. This highlights the potential of protocol-guided generative fusion beyond conventional pixel-level fusion.

\subsection{Additional Nighttime Comparisons}\label{sec:supp_night}
In Fig.~\ref{fig:final_visual}, the nighttime scenes are mainly affected by low illumination, haze-like degradation, light glare, and reduced contrast. 
Although the infrared images provide clear responses for salient objects such as pedestrians and vehicles, directly introducing infrared information may lead to unnatural brightness, color shifts, or excessive thermal responses.

Existing methods often produce washed-out regions around light sources and road surfaces, or weaken structural details in dark and hazy areas. 
In contrast, ProtoFusion maintains a better balance between thermal saliency and visible appearance, preserving important targets while producing cleaner illumination, more stable colors, and clearer scene structures.

\section{Failure Case Analysis}\label{app:failure_analysis}
\begin{figure*}[!t]
\centering
\includegraphics[width=0.322\linewidth,trim=0 177bp 359bp 0,clip]{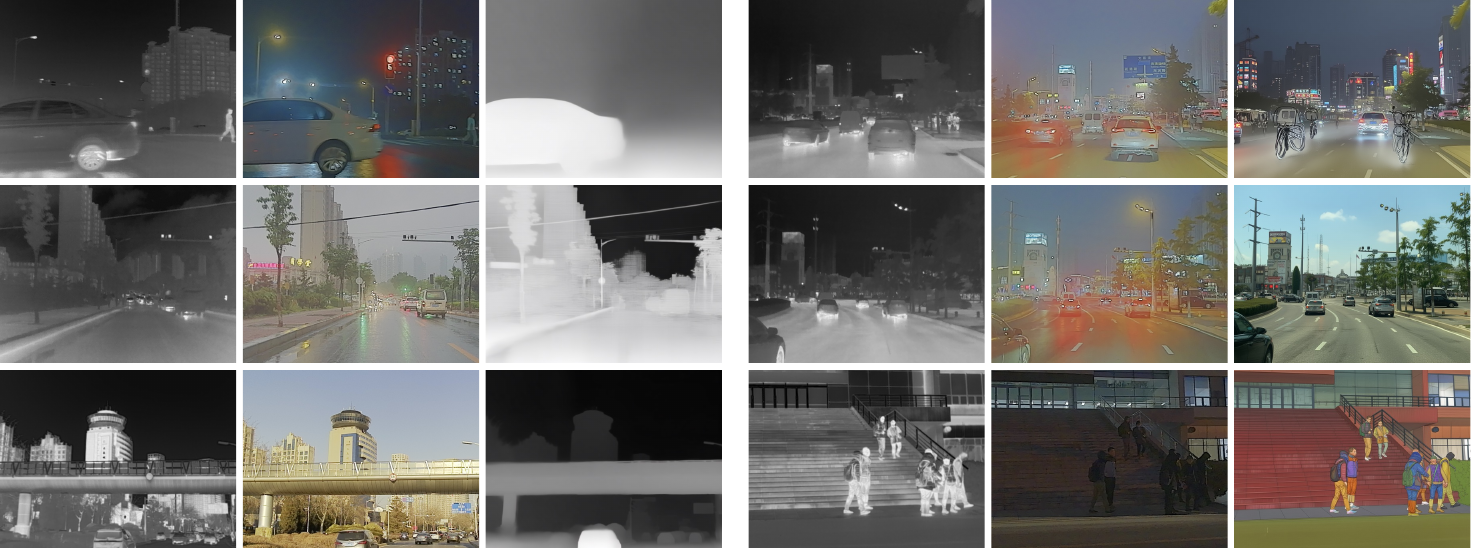}\hfill
\includegraphics[width=0.322\linewidth,trim=359bp 177bp 0 0,clip]{Fig08_fixed_prompt_failure.pdf}\hfill
\includegraphics[width=0.322\linewidth,trim=0 89bp 359bp 88bp,clip]{Fig08_fixed_prompt_failure.pdf}\par\smallskip
\includegraphics[width=0.322\linewidth,trim=359bp 89bp 0 88bp,clip]{Fig08_fixed_prompt_failure.pdf}\hfill
\includegraphics[width=0.322\linewidth,trim=0 0 359bp 177bp,clip]{Fig08_fixed_prompt_failure.pdf}\hfill
\includegraphics[width=0.322\linewidth,trim=359bp 0 0 177bp,clip]{Fig08_fixed_prompt_failure.pdf}
\par\smallskip
\fontsize{8}{9.2}\selectfont
\setlength{\tabcolsep}{0pt}
\renewcommand{\arraystretch}{0.95}
\begin{tabularx}{\linewidth}{*{9}{>{\centering\arraybackslash}X}}
\oneline{Infrared} & \oneline{Visible} & \oneline{Fusion Image} &
\oneline{Infrared} & \oneline{Visible} & \oneline{Fusion Image} &
\oneline{Infrared} & \oneline{Visible} & \oneline{Fusion Image}\\
\end{tabularx}
\caption{Failure cases of fixed prompting for infrared and visible image fusion. 
Although the same handcrafted prompt is used for all image pairs, the generated results show diverse failure modes, including structure collapse, modality domination, loss of visible scene texture, hallucinated objects, unrealistic illumination changes, and severe style drift. 
These failures show that one-shot fixed prompting cannot reliably constrain the large synthesis space of a generative fusion model, motivating protocol-guided prompt refinement.}
\label{fig:fixed_prompt_failure}
\end{figure*}

\subsection{Limitations of Fixed Prompting in Generative Image Fusion}
\label{sec:supp_motivation_prompt_refinement}
A simple way to use a multimodal generative model for infrared and visible image fusion is to provide a fixed, manually designed prompt that asks the model to preserve infrared thermal targets, retain visible textures, and produce a natural fused image. Although this strategy is intuitive, we find that it is not reliable in practice. As shown in Fig.~\ref{fig:fixed_prompt_failure}, the same fixed prompt can produce substantially different failure modes across image pairs. In some cases, the generated image collapses into an over-smoothed or low-contrast result, where visible structures and infrared targets are both weakened. In other cases, the output is dominated by one modality; for example, it may resemble an infrared response map that lacks visible color and scene texture. More seriously, the generative model may change the scene itself, hallucinating unsupported objects, converting a rainy or nighttime scene into a bright daytime scene, or producing cartoon-like colorization and unnatural target rendering.

These examples show that the limitation of fixed prompting lies not only in weak control, but also in its inability to specify image-pair-dependent fusion constraints. Infrared and visible image fusion must simultaneously preserve thermal targets, retain source-consistent visible structures, suppress degradations, and avoid hallucinated content or style drift. However, the dominant failure mode varies across image pairs: infrared evidence may be underused, thermal responses may be over-transferred, visible degradations may leak into the result, or the generative prior may rewrite the scene. Therefore, we treat the prompt as an optimizable interface rather than a fixed instruction. Through protocol-guided prompt refinement, the current result is diagnosed and the structured protocol is updated accordingly, enabling the next prompt to correct concrete failures such as modality imbalance, structure loss, thermal overpainting, and appearance drift while keeping the fusion intent explicit and interpretable.

\begin{figure}[ht]
\centering
\includegraphics[width=0.32\linewidth,trim=0 0 1242bp 0,clip]{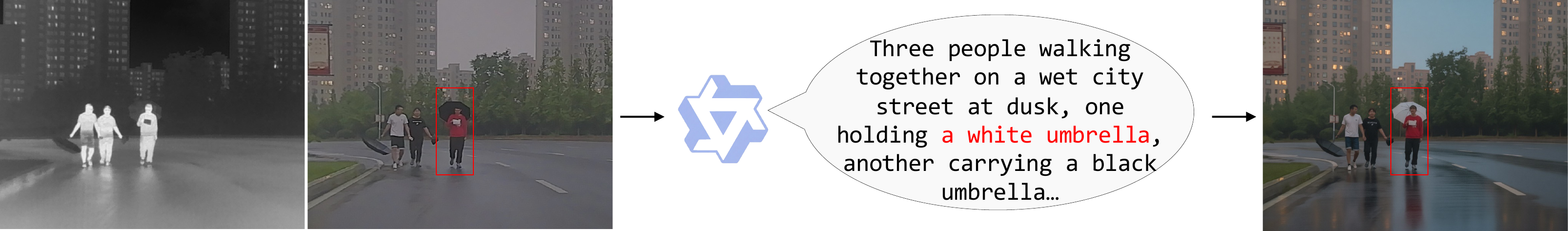}\hfill
\includegraphics[width=0.32\linewidth,trim=302bp 0 939bp 0,clip]{Fig12_Hallucination.pdf}\hfill
\includegraphics[width=0.32\linewidth,trim=1242bp 0 0 0,clip]{Fig12_Hallucination.pdf}
\par\smallskip
\small
\setlength{\tabcolsep}{0pt}
\begin{tabularx}{\linewidth}{*{3}{>{\centering\arraybackslash}X}}
Infrared & Visible & Fused Image\\
\end{tabularx}
\par\smallskip
{\footnotesize
\textbf{VLM interpretation:} ``Three people walking together on a wet city street at dusk, one holding a \textcolor{red}{white umbrella}, another carrying a black umbrella\ldots''\par}
\caption{Qualitative example of hallucination propagation in generative infrared and visible image fusion. A hallucinated ``white umbrella'' in the interpretation of the visible image is propagated into the fused result, producing a source-inconsistent semantic artifact.}
\label{fig:hallucination}
\end{figure}

\subsection{Hallucination in Image Fusion}\label{sec:hallucination_appendix}
ProtoFusion formulates infrared and visible image fusion as protocol-constrained generation, where the fused image is produced under protocol-guided prompt control by a multimodal image generator. This formulation provides strong semantic flexibility, but it also introduces a characteristic failure mode: hallucinated semantics may be propagated into the fusion result. In our current implementation, this risk is intrinsic to the generation pipeline, since the image generator already involves an internal visual understanding stage. As a result, once the internal understanding stage produces an incorrect semantic interpretation, the downstream generator may render this error as if it were valid scene evidence.

Figure~\ref{fig:hallucination} shows a representative failure case. The interpretation of the visible image incorrectly introduces a non-existent ``white umbrella,'' which is then materialized in the fused image as a semantic artifact. This artifact is not supported by the infrared or visible inputs; instead, it arises because the generative fusion process follows a semantically plausible but source-inconsistent cue. The example reveals a key limitation of generative fusion: semantic guidance can improve controllability and interpretability, but erroneous visual understanding may also inject false content into the fused result.

This failure case is consistent with the limitation discussed in the main paper, namely that pretrained multimodal generative models may produce hallucinated structures or source-inconsistent details in difficult cases. For infrared and visible image fusion, this issue is particularly important because the objective is not only visual coherence, but also faithfulness to cross-modal evidence. It therefore motivates stronger source-grounded semantic verification and cross-modal consistency control in future generative fusion systems.

\end{document}